\documentclass[authoryear]{article}

\usepackage[utf8]{inputenc} 	
\usepackage{graphicx}      		
\usepackage[round]{natbib}
\usepackage{amsfonts}        	
\usepackage{amsmath,amssymb}    
\usepackage{bm}					
\usepackage{xcolor}             
\usepackage{float}              
\usepackage[caption=false]{subfig} 
\usepackage{multirow}  			
\usepackage{tikz}				
\usepackage{amsthm}

\usepackage{breakurl}           
\usepackage[breaklinks, hidelinks]{hyperref}  
 \usepackage{authblk}            

\usetikzlibrary{decorations.pathreplacing,calligraphy}

\newtheorem{definition}{Definition}
\newtheorem{theorem}{Theorem}
\newtheorem{proposition}[theorem]{Proposition} 

\newcommand{\figrnn}{
	\begin{tikzpicture}[scale=1.1]
	
	\def\H{0.5}
	\def\D{0.25}
	\def\R{0.125}
	\def\W{1.8}
	
	\node[anchor=south] at (0, 0) {$z^{(k)}$};
	
	\foreach \y/\l in {1.0/{$W_k g(z^{(k)}) + b_k$}, 2.0/{$W_{k+1} g(z^{(k+1)}) + b_{k+1}$}}{
		\draw[thick, ] (-\W,-\y) -- (-\W,-\y-\H) -- (\W,-\y-\H) -- (\W,-\y) -- cycle;
		\node[anchor=center] at (0, -\y-\D) {\l};
	}
	
    \draw [thick, ->] (0, 0)  -- (0, -1);
    \draw [thick, ->] (0, -1.5)  -- (0, -2.0);
    \draw [thick, ->] (0, -2.5)  -- (0, -3+\R);
    \draw [thick, ->] (0, -3-\R)  -- (0, -3.5);
	\node[anchor=north] at (0, -3.5) {$z^{(k+2)}$};
	
	\draw[thick, fill=white](0, -3) circle (\R);
	\node[anchor=center] at (0, -3) {$+$};
	\draw[fill] (0, -0.5) circle (0.05cm);
	\draw[thick, ->] (0, -0.5) -- (-\W-\D, -0.5) -- (-\W-\D, -3) -- (-\R, -3);
	
	\end{tikzpicture}
}

\newcommand{\figachitectures}{
\begin{tikzpicture}[scale=1.0]

\def\XA{2.0}
\def\XB{8.0}
\def\YB{3}

\def\HL{2.5}
\def\HH{1.0}

\foreach \xi/\yi/\li/\hin/\hout in {\XA/1/c/d_x+m/, \XB/1/d/d_x+d_\beta/, \XA/\YB/a/d_x/m, \XB/\YB/b /d_x/d_\beta}{
    \draw (\xi, \yi)   -- (\xi+\HL, \yi) --(\xi+\HL, \yi+\HH) -- (\xi, \yi+\HH)  -- (\xi, \yi); 
    \node[anchor=south east] at (\xi-1, \yi+\HH) {$\li)$};
    \node[anchor=center] at (\xi+0.5*\HL,\yi+0.5*\HH) {$\mathbf{R}^{\hin} \to \mathbf{R}^{\hout}$};
}

\foreach \xi/\yi/\li in {\XA/1/c, \XB/1/\beta}{
    \draw[->] (\xi-1, \yi+0.75) node[anchor=east]{$x_{ij}$}-- (\xi, \yi+0.75);
    \draw[->] (\xi-1, \yi+0.25) node[anchor=east]{$\li_{j}$}-- (\xi, \yi+0.25);
    \draw[->] (\xi+\HL, \yi+0.5) -- (\xi+\HL+1, \yi+0.5)node[anchor=west]{$y_{ij}$};
}

\foreach \xi/\yi/\li in {\XA/\YB/c, \XB/\YB/\beta}{
    \draw[->] (\xi-1, \yi+0.5) node[anchor=east]{$x_{ij}$}-- (\xi, \yi+0.5);
    \draw[->] (\xi+\HL, \yi+0.5) -- (\xi+\HL+0.35, \yi+0.5);
    \draw[->] (\xi+\HL +0.65, \yi+0.5) -- (\xi+\HL+1, \yi+0.5)node[anchor=west]{$y_{ij}$};
    \node[anchor=center] at (\xi+\HL+0.5,\yi+0.5) {$\cdot$};
    \draw[->] (\xi+\HL+0.5,\yi+1.0) node[anchor=south]{$\li_j$} -- (\xi+\HL+0.5,\yi+0.65);
    \draw (\xi+\HL+0.5,\yi+0.5) circle (0.15);
}
	
\end{tikzpicture}
}

\title{Multi-task neural networks by learned contextual inputs}
\author[1,2]{Anders T. Sandnes\footnote{Corresponding author: anders@solutionseeker.no}}

\author[1,3]{Bjarne Grimstad}
\author[2]{Odd Kolbjørnsen}

\affil[1]{Solution Seeker AS, Oslo, Norway}
\affil[2]{Department of Mathematics, University of Oslo, Oslo, Norway}
\affil[3]{Department of Engineering Cybernetics, Norwegian University of Science and Technology, Trondheim, Norway}

\begin{document}
\maketitle

\begin{abstract}
This paper explores learned-context neural networks. It is a multi-task learning architecture based on a fully shared neural network and an augmented input vector containing trainable task parameters. 
The architecture is interesting due to its powerful task adaption mechanism, which facilitates a low-dimensional task parameter space. Theoretically, we show that a scalar task parameter is sufficient for universal approximation of all tasks, which is not necessarily the case for more common architectures. 
Empirically it is shown that, for homogeneous tasks, 
the dimension of the task parameter may vary with the complexity of the tasks,
but a small task parameter space is generally viable.
The task parameter space is found to be well-behaved, which simplifies workflows related to updating models as new data arrives, and learning new tasks with the shared parameters are frozen.
Additionally, the architecture displays robustness towards datasets where tasks have few data points.
The architecture's performance is compared to similar neural network architectures on ten datasets, with competitive results.
\end{abstract}

\section{Introduction} \label{sec:introduction}
A remarkable feat of nature is its ability to create endless variations on concepts that seem to be rigorously defined.
Across all domains, we find classes of objects, clusters of phenomena, and groups of individuals, all of which seem to follow some overarching set of rules.
And still, each separate instance puts a unique spin on the outcome.
While this is fascinating to observe, it can be frustrating to model.
One can quickly run into both performance issues and operational challenges.
An isolated problem may have too few data points to produce a satisfying model. 
Or, there are just too many models to train and maintain for a family of problems.

Multi-task learning, as presented by \citet{Caruana1997}, 
is a learning strategy where multiple related models are trained simultaneously. 
The models may share a subset of their parameters, which allows them to capture general domain knowledge.
The purpose is to improve model generalization, by effectively training the shared parameters on more data.

Model frameworks based on the multi-task philosophy appear in many sciences.
In the statistical literature, they are, among others, referred to as mixed-, hierarchical-, multi-level-, or random-effect models. 
These methods see frequent use in sociology, economics, biometrics, and medicine \citep{demidenko2004, raudenbush2002hierarchical}. 
These models are often of moderate size and complexity.
In the machine learning domain, there is broad diversity within transfer learning and multi-task learning \citep{Lu2015, zhang2021}. 
Methods that combine transfer learning or multi-task learning with deep learning have seen success in several domains, 
with extensive research going into areas such as image analysis \citep{Zamir2018, Kokkinos2017, Morid2021} and natural language processing \citep{raffel2020, devlin2019, brown2020}. 
Engineering domains, such as solar and wind power \citep{Wu2021, Dorado-Moreno2020}, have also seen some multi-task learning applications. 
However, in many cases, research is still needed to create satisfying solutions \citep{curreri2021}.

Our focus is on \textit{non-linear regression} problems. 
We are particularly interested in problems with many similar tasks that have complex input-output relationships. 
The tasks may have limited data. 
This class of problems captures many modeling challenges within engineering and industrial systems. 
Examples are cases with multiple instances of the same object, such as turbines in a wind farm, or batch processes, such as biomass accumulation in a fish farm or a field of crops. 
Here, each task typically has few observations, but they are, by design, quite similar. 
Solving such problems requires an architecture with a high degree of flexibility, where task adaptation can be done with few data points. 
Additionally, the architecture must facilitate the many operations needed to keep machine-learning models running in practice. 
Examples may be model updates in case of time-varying conditions,
or the identification of new tasks that arrive at a later time.

We study \textit{learned-context neural networks}.
The architecture consists of two components.
A feedforward neural network where all parameters are shared, and a set of task-specific parameter vectors.
The task parameter vector serves as additional inputs to the neural network,
that modulates the neural network computation.
They are referred to as a \textit{learned} context because they are found during training simultaneously with the shared neural network.
The learned context is a powerful task adaptation mechanism that can achieve a high degree of task adaptation with few task-specific parameters. 
Additionally, it can facilitate the identification of a well-behaved task parameter space. 
By this, we mean a space that captures continuous latent properties of the tasks themselves, rather than just being a task encoding. 
A well-behaved task parameter space is desirable because it enables us to train the shared network once, and then only be concerned with the task parameter space for day-to-day operations. 
This is especially useful if a complete re-training takes significant manual effort, is computationally expensive, new data points arrive frequently, or access to the shared model is limited \citep{Liu2023promptsurvey,Lester2021power}.

Variations of learned-context neural networks have, to our knowledge, only been applied in a meta-learning study by \citet{zintgraf19a} and a virtual flow metering application by \citet{Sandnes2021}. 
\citet{zintgraf19a} proposed a model-agnostic meta-learning method, which is concerned with context parameters in general. Learned-context neural networks are used as one of the concrete examples in an experimental study. 
\citet{Sandnes2021} use a variation of the architecture to study the benefit of multi-task learning for a soft sensing application. The focus is on the benefits of a multi-task architecture compared to the single-task learners traditionally used within the domain.
The learned-context neural network itself has never been thoroughly analyzed.

\subsection{Contributions}
We provide a deep dive into theoretical and practical aspects of the learned-context neural network architecture. 
We explore the task adaptation mechanism and relate it to existing methodology within statistics and machine learning.
We prove that a scalar task parameter is sufficient for the universal approximation of a set of tasks theoretically.
The viability of such a low dimensional task parameter space is studied empirically for homogeneous tasks, 
and the benefits of a larger task parameter space are explored.
The performance of the learned-context neural networks is compared to similar architectures on ten datasets. 
Comparisons are made on predictive performance, sensitivity to dataset size, and the effect of the task parameter dimension.

\subsection{Outline}
The content of the paper is organized as follows.
Section \ref{sec:background} presents notation and gives an introduction to mixed models, multi-task learning, and other concepts related to the learned-context neural networks.
Section \ref{sec:mmnn} gives a theoretical description of learned-context neural networks,
and Section \ref{sec:universal-approximation} studies its universal approximation capabilities.
Section \ref{sec:results-empirical} presents an empirical analysis of learned-context neural networks and compares it to similar architectures.
Section \ref{sec:discussion} summarizes and discusses the findings,
and Section \ref{sec:conclusion} concludes the paper.

\section{Background} \label{sec:background}

Here we present a minimal background required to study learned-context neural networks.
Starting with a brief discussion on feedforward neural networks, followed by an introduction to multi-task learning.

\subsection{Feedforward neural networks}
\label{sec:background-neural-networks}

The concept of neural networks covers a broad range of model architectures \citep{Schmidhuber2015}.
We focus on feedforward neural networks for regression problems \citep{Goodfellow-et-al-2016}.
Let our features be $x \in \mathbf{R}^{d_x}$ and our labels be $y \in \mathbf{R}^{d_y}$.
A feedforward neural network is then a function $f: \mathbf{R}^{d_x} \to \mathbf{R}^{d_y}$,
composed of $K$ nonlinear transforms,
$f^{(k)}:\mathbf{R}^{d_k} \to \mathbf{R}^{d_{k+1}}$,
known as layers, according to
\begin{align}
    f(x) = (f^{(K)} \circ \dots \circ f^{(1)})(x). \label{eq:neural-net-as-composition}
\end{align}
We refer to the dimension $d_{k+1}$ as the width of the $k$th layer,
and the number of layers, $K$, as the depth of the network.
Each layer is a transform that builds on the output of the previous layer.
The input to the first layer is our features,
and the output of the last layer is our predicted labels.

A common type of layer is to combine an affine mapping, 
parametrized by a matrix $W_k \in \mathbf{R}^{d_{k+1}\times d_k}$ and a vector $b_k \in \mathbf{R}^{d_{k+1}}$,
with an \textit{activation function},
$h_k: \mathbf{R} \to \mathbf{R}$,
\begin{align}
    f^{(k)}(z_k) = h_k(W_k z_k + b_k), \label{eq:nn-layer-specification}
\end{align}
The activation function, $h_k$, is applied \textit{element wise} to its vector argument.
The rectified linear unit, 
$h_k(\cdot) = \max(0, \cdot)$
is a common choice for activation for hidden layers, $k=1, \dots, K-1$,
while the activation on the last layer, $k=K$, depends on the learning problem \citep{Goodfellow-et-al-2016}.
For regression problems, we use the identity function as activation in the last layer.

A neural network constructed with layers such as Equation \ref{eq:nn-layer-specification} is an \textit{universal approximator}.
Universal approximation refers to the ability of neural network architectures to approximate a function with arbitrary precision,
given a set of qualifiers on the function and the network \citep{cybenko89, Hornik1989}.
Several variations of universal approximation properties have been shown for different neural network architectures,
including ReLU-based feedforward networks \citep{Lu2017, kidger20a}. 

While a sufficiently wide and deep feedforward network is a universal approximator,
there are many compelling reasons to consider augmentations to this architecture.
For instance, residual neural networks introduce skip-connections between layers \citep{He2016}.
These connections add the output of an early layer to the input of a later layer.
Skip-connection can improve the training procedure by facilitating a better flow of gradients through the networks \citet{balduzzi2017}.
As an example,
augmenting the feedforward neural network described by
Equation \ref{eq:neural-net-as-composition} and Equation \ref{eq:nn-layer-specification} 
with skip-connections that span two layers, 
would yield a residual neural network with three main elements.
The first linear layer,
\begin{align}
	z^{(2)} = W_1 x + b_1, \label{eq:nn-res-first-linear}
\end{align}
a sequence of residual blocks
\begin{align}
    \left.\begin{aligned}
    z^{(k+1)} &= W_{k} g\left(z^{(k)}\right) + b_{k} \\
	z^{(k+2)} &= z^{(k)} + W_{k+1} g\left( z^{(k+1)}\right)+ b_{k+1} 
  \end{aligned}\right\rbrace
	k = 2, 4, \dots, K-2,\label{eq:nn-res-blocks}
	\end{align}
and a final linear layer
\begin{align}
	y = W_{K} z^{(K)} + b_{K}, \label{eq:nn-res-last-linear}
\end{align}
where the identity activation function is omitted.
The residual blocks in Equation \ref{eq:nn-res-blocks} use a pre-activated structure \citep{He2016}, and are illustrated in Figure \ref{fig:residual-block}.

\begin{figure}
     \centering
	    \figrnn
     \caption{Residual block spanning two hidden layers.
     Each layer consists of an activation function and an affine transformation. 
     The block output is the sum of the block input and the result of applying the two layers to the input.}
     \label{fig:residual-block}
 \end{figure}

Other augmentations to the architecture described above can be to construct layers that capture properties inherent to the data and learning problem. 
Examples are convolutional neural networks
which exploits translational invariance of the input-output relationship \citep{LeCun1989},
or
recurrent neural networks, which sequentially process an input of arbitrary size \citep{Goodfellow-et-al-2016}.
Recently, transformer-based architectures have leveraged the attention mechanism to achieve great results in sequence-to-sequence modeling tasks without relying on recurrence or convolution \citep{Vaswani2017}.

\subsection{Multi-task learning}
\label{sec:background-multitask-learning}
Multi-task learning is a branch of machine learning where multiple learning problems are solved simultaneously, where mechanisms are introduced, either in the model or the learning algorithm,
to facilitate knowledge sharing between the different problems \citep{zhang2021}. 
The idea is that learning problems may have aspects in common,
and that information contained in data from one problem may be beneficial for other problems.
Each learning problem is referred to as a \textit{task}.
We consider a set of $m$ tasks $\left\lbrace (\mathcal{D}_j, f_j) \right\rbrace_{j=1}^{m}$.
Each task consists of a set of observations $\mathcal{D}_{j} = \left\lbrace (x_{ij}, y_{ij}) \right\rbrace_{i=1}^{n_j}$,
where $x_{ij} \in \mathbf{R}^{d_x}$ and $y_{ij} \in \mathbf{R}$, and a function $f_j: \mathbf{R}^{d_x} \mapsto \mathbf{R}$ .
These are related by
\begin{align}
y_{ij} = f_j(x_{ij}) + \epsilon_{ij}, \label{eq:task-function}
\end{align}
where $\epsilon_{ij}$ is a random variable with expected value equal to zero.
Further, let the indicator variable $c_j \in \left\lbrace 0, 1 \right\rbrace^{m}$ be a one-hot task encoding
where the $j$th element is one and the rest are zeros.
We consider \textit{homogeneous} multi-task learning, 
which means that the elements of $x$ and $y$ represent the same quantities across all tasks.
Additionally, we limit our scope to regression problems, which means that the labels, $y$, are continuous quantities.

Multi-task learning attempts to improve the learning of task functions in Equation \ref{eq:task-function} by 
knowledge-sharing mechanisms. 
These mechanisms are often divided into two main categories, \textit{soft parameter sharing} and \textit{hard parameter sharing} \citep{Vandenhende2022}.
In soft parameter sharing, each task has its own parameters, and the knowledge transfer is done by sharing the features each task has learned or through regularization mechanisms that promote similarity between the task parameters.
In hard parameter sharing, there is a set of shared parameters, that are reused by all tasks, in addition to the task-specific parameters.
The number of parameters in total, across all tasks, grows faster with soft parameter sharing because each task must have its own set of all parameters, which can be a concern in some applications \citep{Vandenhende2022}.

We limit our discussion to hard parameter sharing where
task adaptation is achieved using a small set of task parameters.
We use $\alpha$ to denote a set of shared parameters 
and $\beta_j\in \mathbf{R}^{d_\beta}$ to denote a vector of task-specific parameters.
Our general model for the task functions in Equation \ref{eq:task-function} is then
$f_{j}(x_{ij}) = f(x_{ij}; \alpha, \beta_j)$.

The concept of multi-task learning has been studied extensively for linear models under a variety of names and contexts.
The simplest form is a \textit{varying-intercept} linear model,
\begin{align}
    f_j(x_{ij}) = a^\top x_{ij} + b + \beta_j, \label{eq:linear-varying-intercept}
\end{align}
where each task is a linear function of the observed variables and tasks differ only in the individual bias terms $\beta_j$ \citep{Gelman2013}.
With such a setup it is common to assume task parameters are distributed according to $\beta_j \sim \mathcal{N}(0, \sigma_\beta^2)$.
Task-specific slope parameters are also common. 
These models, known as multilevel-, hierarchical-, or mixed effect models, are extensively described in the statistical literature \citep{demidenko2004, raudenbush2002hierarchical}.
An extension to the linear model is to factorize the parameters into a task component and a shared component,
a notable example being the group-and-overlap model of \citet{Kumar2012},
$f_j(x_{ij}) = a_{j}^\top x_{ij} + b_{j}$,
where the parameters are found as
\begin{align}
\begin{bmatrix} a_{j} \\ b_{j}\end{bmatrix} = L \beta_{j}. \label{eq:go-mtl-parameters}
\end{align}
Task-specific parameters are linear combinations of latent basis tasks,
given as the columns of $L$. 
A tailored algorithm is used to jointly optimize the latent task matrix $L$ and the individual task parameters. 
The optimization is controlled by the desired number of latent tasks and their sparsity.
This structure allows the degree of sharing between tasks to be found during training. 
The mechanism introduces robustness towards outlier tasks
because tasks that do not share any aspects with the others can be isolated to separate columns of $L$. 
This allows the main tasks to be differentiated without inference from potential outliers.

Moving beyond linear model structures introduces several design choices.
The simplest is fixed nonlinear analytical expressions where parameters can be partially shared, task-specific, or found through expressions such as Equation \ref{eq:go-mtl-parameters}. 
A classic example is nonlinear growth curve models, 
commonly formulated as logistic curves \citep{demidenko2004}.

Alternatively, the nonlinearities can be selected from a predefined set of candidates as proposed by \citet{Argyriou2008}.
This produces the expression
$f_j(x_{ij}) = \sum_{k=1}^{d_\beta} \beta_{j,k} \phi_k(x_{ij})$,
where the knowledge-sharing mechanism lies in the shared choice of basis functions $\phi_k$.

If the space of basis functions is not predefined,
a more flexible solution is to allow them to be learned by a multi-task neural network \citep{Caruana1997}. 
This can be expressed as 
\begin{align}
	f_j(x_{ij}) = \beta_j^\top h(x_{ij};\alpha), \label{eq:mtl-nn-last-layer-beta}
\end{align}
where $h: \mathbf{R}^{d_x} \to \mathbf{R}^{d_\beta}$ is a shared neural network parametrized by $\alpha$. 
The task parameters $\beta_j$ represent a task-specific output layer.
The multi-task neural network in Equation \ref{eq:mtl-nn-last-layer-beta} is often presented as a fully shared neural network,
$\Tilde{h}: \mathbf{R}^{d_x} \to \mathbf{R}^{m}$,
with one output for each task, in combination with the indicator variable, $c_j$, which selects the output corresponding to the $j$th task,
\begin{align}
	f_j(x_{ij}) = c_j^\top \Tilde{h}(x_{ij};\alpha). \label{eq:mtl-nn-last-layer-c}
\end{align}
These two formulations are identical in practice, but Equation \ref{eq:mtl-nn-last-layer-beta} has the task parameters explicitly stated, while Equation \ref{eq:mtl-nn-last-layer-c} has the task parameters hidden in the last layer of $\Tilde{h}$.

A different neural network strategy is the context-sensitive networks of \citet{Silver2008},
which augments the input to the neural network with the indicator variables, $c_j$,
\begin{align}
	f_j(x_{ij}) &= h(z_{ij};\alpha), \
	z_{ij} = \begin{bmatrix} x_{ij} \\ c_j \end{bmatrix}. \label{eq:mtl-nn-context-sensitive}
\end{align}
All the parameters of the neural network are shared,
but, as with Equation \ref{eq:mtl-nn-last-layer-c},
there are implicit task-specific parameters.
In the context-sensitive network, the task-specific parameters are hidden in the first layer.

\citet{zintgraf19a} introduced context-adaptation, as a meta-learning strategy for fast task adaptation.
The strategy is based on \textit{trainable} model inputs and was tested for several network architectures,
including feedforward neural networks, convolutional neural networks, and reinforcement learning.
The learned-context neural network presented here is equal to a particular instance of the context-adaptation presented by \citet{zintgraf19a},
which is a feedforward neural network similar Equation \ref{eq:mtl-nn-context-sensitive} where the one-hot encoding, $c_j$, is replaced by trainable task parameters.

Other neural network architectures allow all network weights to have some level of task adaptation.
Many of these build on factorizations similar to the method of \citet{Kumar2012}, given in Equation \ref{eq:go-mtl-parameters},
where one factor is shared between tasks and one is task-specific.
\citet{Yang2017} directly extends Equation \ref{eq:go-mtl-parameters} to multidimensional parameters, such as matrices, 
through tensor factorization. 
This allows the method to be used for both the weights, $W_k$, and biases, $b_k$, in the fully connected layer in Equation \ref{eq:nn-layer-specification}.

\citet{li2018measuring} presents another factorization based strategy,
where arbitrary parameters, $\theta^{(D)} \in \mathbf{R}^D$,
are found as 
\begin{align}
\theta^{(D)}  = \theta_0 + P \theta^{(d)}, \label{eq:intrinsic-dimension}   
\end{align}
where $\theta_0 \in \mathbf{R}^D$ and $P \in \mathbf{R}^{D\times d}$ are randomly generated and frozen,
and the only adaptation of the parameters $\theta^{(D)}$ is through $\theta^{(d)} \in \mathbf{R}^d$.
The dimension $d$ of the trainable parameters is taken significantly smaller than the dimension $D$ of the model parameter space.
\citet{aghajanyan-etal-2021-intrinsic} augments Equation \ref{eq:intrinsic-dimension} to deep neural networks,
and use the weights of a pre-trained neural network in place of $\theta_0$.
They then use the low-dimensional parameters, $\theta^{(d)}$, to fine-tune the pre-trained model for new tasks.
The experiments of \citet{li2018measuring} and \citet{aghajanyan-etal-2021-intrinsic} illustrate that for many learning problems, it is possible to achieve high performance by tuning a low-rank adjustment of the model parameter space.

\citet{Mallya2018} suggests masked networks as an alternative strategy for tuning a pre-trained neural network.
If $W_k^\star$ is the pre-trained neural network weights for a layer, such as in Equation \ref{eq:nn-layer-specification},
and $M_k$ is a binary task-specific masking matrix of the same dimension as $W_k^\star$, 
then the task-adapted weights of a masked neural network are given by 
\begin{align}
W_k = W_k^\star \circ M_k, \label{eq:masked-network}
\end{align}
where the multiplication is done elementwise.
In contrast to the low-rank adjustment in Equation \ref{eq:intrinsic-dimension}, the mechanism in Equation \ref{eq:masked-network} requires the number of task-specific parameters to be equal to the number of parameters being tuned.
\cite{Wen2020BatchEnsemble} and \cite{Wang2023} presents a variation of Equation \ref{eq:masked-network},
where the binary matrix, $M_k$,
is replaced by a trainable rank-one matrix,
which yields a different type of low-rank approximation than that of Equation \ref{eq:intrinsic-dimension}.

As an alternative to factorization strategies, one can introduce layers that facilitate sharing between otherwise disjoint neural networks \citep{Misra2016},
or learn the extent of sharing during training by allowing the tasks to branch away from a shared neural network base \citep{guo2020}.

Multi-task learning was originally presented in combination with feedforward neural networks with fully connected layers \citep{Caruana1997},
but recent works have utilized multi-task learning in combination with several different neural network types, for instance, attention-based and convolutional layers \citep{Zhang2022attentiontomography,Han2023dualadaptive}.
The task-adaptation mechanisms discussed above are not tied to any particular type of neural network, but for the remainder of this paper, we limit our treatment to fully connected feedforward neural networks.

\subsection{Related learning paradigms} \label{sec:learning-paradigms}

Multi-task learning resembles other learning paradigms, in particular \textit{transfer learning} and \textit{meta-learning}.
The distinctions between these can be difficult, and their nuances have changed over time.
We base our discussion on the recent survey of \cite{Hospedales2022}.

Transfer learning attempts to improve the performance of a task using information from related source tasks. 
This can, for instance, be achieved by copying an existing model trained on the source tasks and applying optional fine-tuning steps to the parameters. 
This tuning has no concern for the performance of the source tasks.
In contrast, multi-task learning attempts to jointly optimize the performance of all tasks.

Meta-learning also attempts to jointly optimize all tasks, but the objective of the optimization is different.
Multi-task learning is concerned with a fixed set of provided tasks and produces a single joint model.
Meta-learning, on the other hand, attempts to improve learning for the whole distribution of tasks, which can include unseen future tasks.
The output is a \textit{learning procedure} that can be applied to all tasks in the task distribution.
As such, meta-learning can draw parallels to \textit{hyper-parameter optimization}.
However, hyper-parameter optimization only considers performance on the current learning problem,
while meta-learning considers the performance over a family of learning problems that may occur.

Prompt-based learning is a method that originates from natural language processing,
where a frozen pre-trained general-purpose model is adapted to new tasks by augmenting the model input as opposed to tuning the model parameters \citep{Liu2023promptsurvey}.
The motivation for prompt methods is that the models used are so large and resource-demanding to train that it is infeasible to create a new model for each type of application \citep{gu2023vision}.
A common class of prompt design is to prepend natural language instructions and hints to the original task input, and use this to generate the model output \citep{li2021prefixtuning}. 
As opposed to crafting human-readable prompts,
\textit{prompt-tuning} formulate the prompt as a set of task-specific parameters in the model embedding space, which allows them to be found by gradient descent methods \citep{gu2023vision,Lester2021power}. 
An important distinction between the multi-task learning presented in Section \ref{sec:background-multitask-learning} and prompt-tuning is how the shared model is trained.
For prompt-tuning, the shared model has been pre-trained without considering the type of tasks it will be used for.
While multi-task learning models are found by supervised learning using task-specific data.

Special cases of these paradigms may be closely linked in practice.
\citet{Grant2018} connects a particular meta-learning algorithm to hierarchical Bayesian models and hyper-parameter optimization.
\citet{zintgraf19a} develops this meta-learning algorithm further with context-adaptation. 
The multi-task learned-context neural networks studied in this paper are equal to one of the meta-learning context-adaptation mechanisms.

\section{Learned-context neural networks} \label{sec:mmnn}

Learned-context neural networks is a multi-task neural network architecture, 
that augments the inputs to a feedforward neural network with trainable task parameters.
Recall our multi-task regression problem presented in Equation \ref{eq:task-function},
where the features are given as $x_{ij} \in \mathbf{R}^{d_x}$ and the labels as $x_{ij} \in \mathbf{R}$,
for tasks $j = 1, \dots, m$, each with data points $i = 1, \dots, n_j$.
Let $\beta_j \in \mathbf{R}^{d_\beta}$ be the task parameters, 
and $h: \mathbf{R}^{d_x + d_\beta} \to \mathbf{R}$ be a fully shared feedforward neural network with the parameters collected in a set $\alpha$.
We then find the task models in Equation \ref{eq:task-function} as
\begin{align}
    f_j(x_{ij}) &= h(z_{ij}; \alpha), \label{eq:mtl-nn-learned-context}
\end{align}
where the input, $z_{ij}$, is constructed from the features, $x_{ij}$, and the learned context, $\beta_j$,
\begin{align}
	z_{ij} &=  \begin{bmatrix} x_{ij} \\ \beta_j \end{bmatrix}. \label{eq:input-x-beta}
\end{align}

\subsection{Learned-context input yield a task-specific bias}

We now explore the task adaptation mechanism in the learned-context neural network,
and relate it to linear varying intercept models and context-sensitive networks.
Let the first layer feedforward neural network in Equation \ref{eq:mtl-nn-learned-context} use and affine transform as in Equation \ref{eq:nn-layer-specification}.
Using the notation from Equation \ref{eq:nn-res-first-linear},
we get the affine transform 
\begin{align}
 z_{ij}^{(2)} = W_1 z_{ij} + b_1,   \label{eq:learned-context-first-layer-W}
\end{align}
where $z_{ij}$ is the augmented input  given in Equation \ref{eq:input-x-beta}. 
Let $d_2$ be the width of the first layer, making the weight matrix $W_1 \in \mathbf{R}^{d_2 \times (d_x + d_\beta)}$.
Partitioning the weight matrix according to $W_1 = \begin{bmatrix} A & L \end{bmatrix}$,
with $A \in \mathbf{R}^{d_2 \times d_x}$ 
and $L \in \mathbf{R}^{d_2 \times d_\beta}$,
allows us to rewrite Equation \ref{eq:learned-context-first-layer-W} as 
\begin{align}
	z_{ij}^{(2)} &= A x_{ij} + L \beta_j + b_1. \label{eq:context-as-mixed-intercept}
\end{align}
This is recognized as the linear varying-intercept model from Equation \ref{eq:linear-varying-intercept},
where $b_1$ is the population average intercept and $\Tilde{\beta}_j = L\beta_j$ is the individual intercept.
However, linear varying-intercept models are usually concerned with scalar, or low-dimensional, outputs.
The dimension $d_2$ of $z_{ij}^{(2)}$ may be high,
and the term $L\beta_j$ allows for a low-rank approximation of the task intercept.
This is the same mechanism utilized in the group-and-overlap method given in Equation \ref{eq:go-mtl-parameters},
and the method inherits the benefits this provides in terms of outlier robustness.

The learned context parameter in Equation \ref{eq:input-x-beta} resembles the context-sensitive neural network given in Equation \ref{eq:mtl-nn-learned-context}.
The difference is that the context-sensitive networks augment the input with the fixed one-hot encoding of the task number, $c_j$.
This yields an expression similar to Equation \ref{eq:context-as-mixed-intercept},
but with the task intercept, $L\beta_j$, replaced by $L^{(cs)} c_j$, where $L^{(cs)} \in \mathbf{R}^{d_2 \times m}$.
For any task $j$, $c_j$ selects the $j$th column of $L^{(cs)}$.
This essentially makes each task responsible for a number of parameters equal to the layer size, $d_2$,
which can be a significant number even for moderately sized networks.
Reducing the number of these implicit task parameters would also reduce the width of the layer,
which can cause an undesirable bottleneck effect \cite{serra2018}.

There is no inherent limitation that forces the contextual parameters to be used as inputs to the first layer. 
Indeed, \citet{zintgraf19a} suggested that they could be used to augment any layer.
Such extensions could be motivated by the need to condition deeper layers on the contextual inputs.
We suggest the use of residual neural networks to achieve such conditioning. 
For instance, using the architecture described by Equations \ref{eq:nn-res-first-linear}--\ref{eq:nn-res-last-linear},
the skip-connections will provide deeper layers with access to the learned-context inputs.

\subsection{Comparable architectures}
Throughout the analysis of the learned-context neural network,
we will make comparisons with two closely related architectures. 
The first is the context-sensitive networks as presented by \citet{Silver2008}, given in Equation \ref{eq:mtl-nn-context-sensitive}.
The second is the classic multi-task neural network of \citet{Caruana1997}, which we will refer to as \textit{last-layer} neural networks.
This is described by Equation \ref{eq:mtl-nn-last-layer-beta}.
The three architectures are illustrated in Figure \ref{fig:mt-architectures}.

\begin{figure}
     \centering
	    \figachitectures
     \caption{
    Illustration of different multi-task neural network strategies. 
    Each box represents a neural network with the dimensions of the domain and co-domain as indicated.
    The neural networks are shared between tasks.
    The circles represent the dot product operation.
    Illustrations $a$ and $b$ are two different views of \textit{the same} architecture,
    namely the last-layer neural network of \citet{Caruana1997}, 
    given in Equation \ref{eq:mtl-nn-last-layer-c} and Equation \ref{eq:mtl-nn-last-layer-beta}.
    Illustration $c$ is the context-sensitive neural network of \citet{Silver2008},
    given in Equation \ref{eq:mtl-nn-context-sensitive}.
    Illustration $d$ is the learned context neural network given in Equation \ref{eq:mtl-nn-learned-context}.
     }
     \label{fig:mt-architectures}
 \end{figure}

\section{Universal task approximation} \label{sec:universal-approximation}

Equation \ref{eq:context-as-mixed-intercept} illustrates how the task adaptation mechanism of learned context neural networks can be viewed as a task-specific bias in the first layer.
In this section, we explore the capabilities of this deceptively simple task adaptation mechanism,
and compare it to the properties of context-sensitive networks and last-layer neural networks.
In particular,
we prove that learned context neural networks possess a multi-task equivalent to the universal approximation property. 
Briefly summarized, 
the main result is that any task adaption is achievable by learned-context neural networks regardless of how many task parameters are used, provided that the shared neural network is sufficiently wide and deep.

We adapt the definitions from \citep{kidger20a} and rely upon their main result for universal approximation.
Our interest is to study the number of task parameters required, so the dimensions of the neural networks themselves are omitted. 
\begin{definition}
Let $\mathcal{F}_{k, l}$ be the class of functions described by feedforward neural networks with input dimension $k$,
output dimension $l$,
an arbitrary number of hidden layers of arbitrary size,
where the ReLU activation function is applied to all hidden units.
\end{definition}
\begin{definition}
Let $C(K)$ be the space of continuous functions $f: K \to \mathbf{R}$, with domain $K \subseteq \mathbf{R}^{d_x}$, where $K$ is compact.
\end{definition}

Consider $m$ tasks given by $y = f_j(x),\ j = 1,\dots, m$,
where $f_j \in C(K)$.
Individually, the functions $f_j$ can be approximated arbitrarily close by a sufficiently sized neural network $\hat{f_j} \in \mathcal{F}_{d_x, 1}$ \citep{kidger20a}.
Extending this notion to multi-task models
means their ability to approximate the composed function 
\begin{align}
    y = f(x, j) = \sum_{k=1}^m I(j=k) f_k(x). \label{eq:task-function-composition}
\end{align}
Here, $I(j=k)$ is the indicator function, tanking the value one if $j=k$ and zero otherwise.

In Equation \ref{eq:task-function-composition}, the inputs of $f(x, j)$ closely resemble those of context-sensitive networks,
making this a natural starting point.
\begin{proposition}
\label{prop:cs}
There exists a context-sensitive neural network from the class $\mathcal{F}_{d_x+m, 1}$
that is arbitrarily close to the multi-task function $f$ of Equation \ref{eq:task-function-composition} with respect to the uniform norm.
\end{proposition}
\begin{proof}
An equivalent computation to Equation \ref{eq:task-function-composition} is
$f(x, c_j) = \sum_{k=1}^m c_{j,k} f_k(x)$, where $c_{j,k}$ is the $k$th element of $c_j$.
Relaxing the indicator variable domain to $c_j \in [0, 1]^m$,
allows the context-sensitive input vector, $z$, to exist in a compact subspace $K^{+} \subseteq \mathbf{R}^{d_x+m}$.
The relaxation gives $f \in C(K^{+})$, a space of which the class $\mathcal{F}_{d_x+m, 1}$ is dense with respect to the uniform norm.
It follows that context-sensitive networks can approximate the set of task functions arbitrarily well.
\end{proof}
This result means that the context-sensitive architecture is sufficiently flexible to achieve the same approximation power as using an individual neural network for each task.  However, it does not say anything about the number of parameters required to achieve this.

Learned-context neural networks can achieve the same universal approximation power as context-sensitive networks, using only a scalar task parameter.
\begin{theorem}
\label{thm:ua}
There exists a learned-context neural network from the class $\mathcal{F}_{d_x+1, 1}$
and a set of task parameters $\beta_j \in \mathbf{R}, \ j = 1, \dots, m$,
that is arbitrarily close to the multi-task function $f$ of Equation \ref{eq:task-function-composition} with respect to the uniform norm.
\end{theorem}
\begin{proof}
The proof is to construct the mapping $(x, \beta_j) \mapsto (x, c_j)$, as the first two hidden layers of a neural network.
The remainder of the network could then be taken as a context-sensitive network.

First, we construct a similar triangle wave as in Figure \ref{fig:pyramid} for the task parameters.
Let the task parameters be assigned values $\beta_j = j$, which is a trivial task encoding.
The first layer is assigned the weights
\begin{align*}
 W_1 &= 1_{2m}, \
 b_1^\top = -\begin{bmatrix} 1-\delta &  1 & 2-\delta &  2 & \dots & m-\delta & m \end{bmatrix},
\end{align*}
where $1_{2m}$ is a vector of $2m$ ones and $\delta \in (0, 0.5)$ is a number determining the slope and spacing of the triangles. 
After ReLU activation, this gives a shifted sequence of the task parameter.
The second layer is assigned the weights
\begin{align*}
 W_2 &= \frac{1}{\delta}  I_m \otimes \begin{bmatrix} 1 & -2 \end{bmatrix}, \ 
 b_2 = 0_m
\end{align*}
where $\otimes$ denotes the Kronecker product and $0_m$ is a vector of $m$ zeros.
This leads to the $j$th entry of $g(z^{(3)})$ to be given by
\begin{align}
    g(z_{j}^{(3)}) = 
    \begin{cases}
    (\beta - j + \delta)/\delta & \text{if } \beta \in [j-\delta, j), \\
    (-\beta + j +\delta)/\delta  & \text{if } \beta \in [j, j+\delta), \\
    0  & \text{otherwise }.
    \end{cases} \label{eq:beta-pyramid}
\end{align}
Only one of the entries of $g(z_{j}^{(3)})$ will be non-zero at a time,
and when evaluated at $\beta = j$ the $j$th entry will be one and the rest zero. 
The sharpness of the transition between zero and one is adjusted by $\delta$.
The weights ($W_1$, $b_1$, $W_2$, and $b_2$) can now be augmented to store a copy of the other input values $x$,
which makes the hidden layer $z^{(3)}$ the same as the input layer of a context-sensitive network.
Therefore,
a learned-context network with a scalar task parameter share approximation properties with the context-sensitive networks from Proposition \ref{prop:cs}.
\end{proof}
This result only considers the number of task parameters and does not put any bounds on the number of shared parameters. While a scalar task parameter is theoretically sufficient, it does not indicate that this is the ideal choice. Indeed, the proof is based on the scalar task parameter taking an indicator role, which is counterproductive to our desire to learn the latent properties of the tasks. This is useful to bear in mind when selecting hyperparameters, as too few task parameters may force an indicator behavior.

The last-layer neural network is less flexible in its task adaptation mechanism. As such, it may require more task parameters than the learned-context networks.
\begin{proposition}
\label{prop:ll}
Last-layer neural networks with base network from the class $\mathcal{F}_{d_x, k}$ and $\beta_j \in \mathbf{R}^k$
require $k \geq m$ to guarantee the existence of a function arbitrarily close to the multi-task function $f$ in Equation \ref{eq:task-function-composition}.
\end{proposition}
\begin{proof}
The last-layer network is a linear combination of $n_\beta$ basis functions,
$y = \beta_j^\top h(x)$.
Let the tasks be given as $m$ sine waves $y = \sin (\omega_j x)$, 
with frequency $\omega_j = j$ and $x \in [-\pi, \pi]$.
These cannot be constructed by a set of basis functions of dimension lower than $m$,
because they are orthogonal to each other.
Hence, in this case, the shared neural network cannot have an output dimension less than $m$.
\end{proof}
Task adaption in the last-layer neural network is through a linear combination of features produced by the fully shared neural network. 
This means that it is limited to scaling the given nonlinearities. 
Contrary, in the learned-context architecture the task parameters can influence the nonlinearities themselves, which leads to a much broader range of adaptions it can achieve.

We note that, as for all neural networks in general, 
there is a difference between the theoretical approximative power of a neural network and the expressivity that is seen in practice
\citep{Jiang2023}.
In particular, the problem of dying neurons with activation functions such as ReLU can significantly reduce the neural networks' ability to effectively approximate the desired functions. 
For the learned-context neural network, the combination of dying neurons and a choice of low dimensional task parameter vector can lead to poor performance in practice. 
We, therefore, advise using a sufficiently wide first layer and monitoring the training properly. 
An empirical study of the viability of low dimensional, and scalar, task parameters is given in the next section.

\section{Empirical analysis}\label{sec:results-empirical}
This section investigates the properties of learned-context neural networks empirically. 
First, the learned-context neural network is compared to similar architectures. The architectures are evaluated on predictive performance, training robustness, sensitivity to the number of data points, and the effect of the number of task parameters. The task parameter space produced by learned-context neural networks is then explored by estimating task parameters for new tasks after the shared parameters are trained and fixed.

\subsection{Benchmark models}
Learned-context neural networks, described in Section \ref{sec:mmnn}, 
are compared to last-layer multi-task networks, described by Equation \ref{eq:mtl-nn-last-layer-beta},
and context-sensitive networks, described by Equation \ref{eq:mtl-nn-learned-context}.
All three network models use the residual structure described in Equations \ref{eq:nn-res-first-linear}--\ref{eq:nn-res-last-linear}.

A linear mixed-effect model acts as a baseline for performance in the multi-task setting. 
The linear models have a set of shared slope and intercept parameters. 
Additionally, each task has its own intercept.
This structure is given in Equation \ref{eq:linear-varying-intercept}.
Discrete features are one-hot encoded for this model.

Additionally,
two \textit{fully shared} models, a neural network and a linear regression model, are used to illustrate the benefit of multi-task learning on the lesser known datasets. Both of these models are trained on data from all tasks, using the features and labels as normal, but ignore the task aspect.

\subsection{Datasets} \label{sec:datasets}
The models are compared on ten datasets.
Two of the datasets are synthetically created to highlight differences between the architectures.
Three datasets, Schools, Parkinson, and Sarcos, are common to the multi-task learning literature \citep{zhang2021},
while the last five are selected from a range of open access data sources.
All datasets correspond to regression problems with a scalar response variable.
The datasets are described below and summarized in Table \ref{tab:datasets}.

\textbf{Frequency} is a synthetic dataset where tasks are sine waves of different frequencies.
Data is generated according to $y_{ij} = 0.5\sin (2\pi \omega_j x_{ij}) + 0.5 + \epsilon_{ij}$.
The task frequencies are sampled as $\omega_j \sim \text{Uniform}(0.5, 4.0)$.
We take $x_{ij}\sim \text{Uniform}(0, 1)$, 
and $\epsilon_{ij} \sim \mathcal{N}(0, \sigma^2)$, $\sigma = 0.1$.

\textbf{Sine and line} is a synthetic dataset proposed by \citet{Finn2018}.
Tasks are taken from two classes,
affine functions, $y_{ij} = a_{j}x_{ij} + b_{j} + \epsilon_{ij}$ 
or sine waves, $y_{ij} = c_{j} \sin(x_{ij} + d_{j}) + \epsilon_{ij}$.  
We generate an equal number of tasks from each class, with parameters sampled as
$a_j, b_j \sim \text{Uniform}(-3, 3)$, 
$c_j \sim \text{Uniform}(0.1, 5.0)$, and
$d_j \sim \text{Uniform}(0, \pi)$. 
For both task types we use $\epsilon_{ij} \sim \mathcal{N}(0, \sigma^2)$, $\sigma = 0.3$,
and $x_{ij}\sim \text{Uniform}(-5, 5)$.
We note that this problem can be represented as linear regression, $y = \beta^\top z $, 
on a set of nonlinear basis functions $z^\top = \begin{bmatrix} 1 &  x & \sin(x) & \cos(x) \end{bmatrix}$
by applying the trigonometric identity $\sin(x + d) = \sin(d)\cos(x) + \cos(d)\sin(x)$.

\textbf{Schools} is a dataset of examination results for students from different schools over a three year period \citep{NUTTALL1989}. 
The goal is to map features of the schools and students to student performance, in order to study school effectiveness.
We treat each school as a task.
The dataset is provided by \citet{schools}.

\textbf{Parkinson} telemonitoring relates patient observations to a disease symptom score \citep{Tsanas2010}. 
Each patient is considered a task.
It is provided by \citet{parkinson}.

\textbf{Bike sharing} relates weather and calendar features to the number of trips of a bike sharing system over two years \citep{Fanaee-T2014}.
Each month is considered a task,
which allows tasks to capture seasonal effects and potential changes to the bike sharing system itself.
Data is provided by \citet{bike}.

\textbf{Inflation} is a dataset from \citet{oecd}. 
It considers the development of the Consumer Price Index (CPI) in 45 countries. 
CPI is taken quarterly from 1970 to 2020, normalized with 2015 being equal to 100\% for each country. 
Each country is a task. Time is the only input variable.

\textbf{Obesity} is a dataset that describes the development of the mean body-mass index from 1985 to 2019 in 200 countries  \citep{Rodriguez-Martinez2020}.
Input variables are age, sex, and year.
The response variable is the percentage considered to be obese.
Each country is a task. 
Data is provided by \citet{ncd-obesity}.

\textbf{Height} is a dataset with the same structure as Obesity,
but the response variable is the mean height within the groups. 
Data is provided by  \citet{ncd-height}.

\textbf{Death rate} describes the rate of death in different age groups.
Input variables are age and sex.
Each pair of year and country is considered a task. There are 183 countries and five years.
This results in a greater number of tasks than Obesity and Height, but fewer input variables.
Data is provided by  \citet{who-deathrate}

\textbf{Sarcos} is data from a robotic arm with seven joints. The goal is to map joint position, velocity, and acceleration to motor torque \citep{Vijayakumar2000}.
Each joint is one task, taking all 21 joint measurements as input to predict torque at the joint. As a consequence, each task has exactly the same input data, differing only in their response.
It is hosted by \citet{sarcos}.

\begin{table*}
\centering
\caption{
Summary of dataset properties. 
Listed are the number of features in the dataset, the number of tasks, the number of data points used for model training and testing,
and the standard deviation of the response variable $y$.
}
\label{tab:datasets}
\begin{tabular}{lrrrrr}
\hline
 Name          &   Num. feat. &   Num. tasks &   Data train &   Data test &   Std. $y$ \\
\hline
 Frequency     &            1 &     250 &   30000 &  25000 &         0.36 \\
 Sine and line &            1 &     100 &    6000 &  10000 &         3.93 \\
 Schools       &            7 &     139 &   12339 &   3023 &        12.72 \\
 Parkinson     &           16 &      42 &    4717 &   1158 &        10.70 \\
 Bike sharing  &            6 &      24 &   13915 &   3464 &       181.38 \\
 Inflation     &            1 &      45 &    6684 &   1344 &        34.61 \\
 Obesity       &            3 &     200 &  126000 & 126000 &         5.22 \\
 Height        &            3 &     200 &  105000 & 105000 &        19.64 \\
 Death rate    &            2 &     915 &   16470 &  16470 &         0.12 \\
 Sarcos        &           21 &       7 &  311388 &  31143 &        18.84 \\
\hline
\end{tabular} 
\end{table*}

\subsection{Training Procedure}
Model parameters are found by minimizing a loss function of mean squared prediction error and parameter regularization \citep{hastie2009elements},
\begin{align}
   \underset{\alpha, \beta_1, \dots, \beta_m}{\min} \ 
   \frac{1}{n}\sum_{j=1}^{m} \sum_{i=1}^{n_j} \left( y_{ij} - f(x_{ij}; \beta_j, \alpha) \right)^2 +
   \lambda_\alpha l(\alpha) +
   \lambda_\beta \sum_{j=1}^{m} l(\beta_j).
        \label{eq:loss-train}
\end{align}
The prediction error is divided by the total number of data points $n$.
Parameters are regularized by the L2 norm,
scaled by factors $\lambda_\alpha$ and $\lambda_\beta$ for shared and task parameters respectively.
The task parameter term is ignored for context-sensitive networks.
All data points from all tasks are weighted equally.
More complex loss functions with individual weights for each task could potentially improve performance in some cases
\citep{Kendall2018, Gong2019},
but this is not considered here.

Optimization of Equation \ref{eq:loss-train} is done by stochastic gradient decent \citep{Bottou2018}, 
with a two-stage learning rate schedule.
Hyperparameters are optimized by the global optimizer LIPO \citep{Malherbe2017}, which is run for a predetermined number of iterations. 
Further details are given in \ref{app:optimizer} for the training procedure 
and \ref{app:hps} for hyperparameters.

\subsection{Base performance}\label{sec:results-base}
The three neural network architectures and the linear mixed model are first compared in terms of test set errors.
The results are summarized in Table \ref{tab:error-base}. 
The neural network models have similar performance on all datasets,
with learned-context and last-layer neural networks being slightly favored over context-sensitive networks in general.
In some cases, such as Obesity and Height, 
we see that the context-sensitive architecture has close to twice the error of the other architectures. 
While this difference is significant, we note that when the errors are compared to the standard deviation of the response variables, given in Table \ref{tab:datasets}, these errors are all quite small.
All methods perform similarly on the Schools dataset, and none were able to explain more than 35\% of the variance.
This is similar to the performance seen in other multi-task learning studies using this dataset \citep{Bakker2003, Evgeniou2005}.
The linear mixed model performed well on two of the datasets, Schools and Parkinson.
In these cases, the neural network models are able to achieve similar performance, which is reassuring.

In theory, given the same dimensions on the neural network, 
the context-sensitive neural network may be expected to be a richer model than the learned-context equivalent.
However, when evaluated on an independent test set, we see that the learned-context neural networks generalize better on most of our datasets,
which may be attributed to their ability to adjust the task parameter dimension and network width independently.

The learned-context and last-layer neural networks are evenly matched, 
being the top performer in four and five out of the ten datasets respectively.
The datasets with the largest relative performance difference between these two models are Death rate, Obesity, Inflation, and Bike sharing, where learned-context networks perform better on the first three and last-layer networks are best on the Bike sharing case. 
However, when comparing the errors reported in Table \ref{tab:error-base} with the dataset standard deviations in Table \ref{tab:datasets}, 
we see that these differences are small in terms of explained variance.
In general, the preferred model will depend on the problem at hand.
For instance, on the synthetic datasets,
the learned-context network performs better on the Frequency data,
and the last-layer neural network performs better on the Sine and line data, which is expected because the architectures match the data-generating processes.

The results in Table \ref{tab:error-base} also include the two fully shared models.
For all datasets, it is beneficial to include the multi-task aspect,
and for many cases, the linear varying intercept model performs better than a fully shared neural network.
The shared models also highlight that some datasets have significant nonlinearities that are shared between all tasks, 
for instance for the Bike sharing, Height, and Death rate datasets.

\begin{table*}
\centering
\caption{
Root mean squared error on test data for all models.
The best score on each dataset is highlighted.
The column headers are 
learned-context neural networks (LC),
context-sensitive neural networks (CS),
last-layer neural networks (LL),
and linear mixed effect models (LME) for multi-task learners,
and linear regression (LR) and fully shared neural networks (FS) for fully shared learners. 
}
\label{tab:error-base}
\begin{tabular}{lllll|ll}
\hline
 Dataset       & LC             &     CS & LL              & LME             &      LR &     FS \\
\hline
 Frequency     & \textbf{0.106} &  0.136 & 0.122           & 0.360           &   0.361 &  0.338 \\
 Sine and line & 0.325          &  0.34  & \textbf{0.316}  & 3.843           &   3.975 &  3.941 \\
 Schools       & 10.203         & 10.363 & 10.314          & \textbf{10.108} &  10.363 & 10.339 \\
 Parkinson     & 2.914          &  2.856 & \textbf{2.670}  & 2.776           &  10.142 &  8.567 \\
 Bike sharing  & 53.869         & 80.208 & \textbf{45.043} & 142.087         & 150.744 & 76.522 \\
 Inflation     & \textbf{1.833} &  2.526 & 2.501           & 11.095          &  18.132 & 17.909 \\
 Obesity       & \textbf{0.123} &  0.281 & 0.210           & 2.512           &   4.522 &  4.401 \\
 Height        & 0.394          &  0.61  & \textbf{0.347}  & 5.055           &   6.763 &  4.735 \\
 Death rate    & \textbf{0.011} &  0.017 & 0.026           & 0.078           &   0.081 &  0.042 \\
 Sarcos        & \textbf{2.176} &  2.188 & 2.482           & 10.653          &  18.293 & 18.275 \\
\hline
\end{tabular}
\end{table*}

To compare the robustness of the training procedure we re-run the training of all neural network models. 
We reuse the hyperparameters and run the training nine additional times.
The best and worst RMSE values for the ten runs combined are given in Table \ref{tab:rmse-variation}.
It also lists the number of times training runs diverged and had to be restarted.
Overall the results are quite consistent.
However, there are cases where the training appears less stable. 
An example is the last-layer neural network on the Height dataset, which yields a large span in relative errors.
These errors are small in absolute value, which makes such comparisons sensitive to the randomness in the training procedure.

\begin{table*}
\centering
\caption{
Results from ten repeated training runs with the same hyperparameter settings for
learned-context neural networks (LC),
context-sensitive neural networks (CS),
and last-layer neural networks (LL).
Reported are the min and max relative RMSE values in each case. 
The values are normalized by the performance of the learned-context neural network in Table \ref{tab:error-base}.
The number of diverging training runs, if any, are given in brackets.
}
\label{tab:rmse-variation}
\begin{tabular}{llll}
\hline
 Dataset       & LC             & CS             & LL             \\
\hline
 Frequency     & 1.00, 1.01     & 1.10, 1.29 (1) & 1.16, 1.21     \\
 Sine and line & 1.00, 1.00     & 1.04, 1.06     & 0.97, 0.97     \\
 Schools       & 1.00, 1.00     & 1.01, 1.02     & 1.01, 1.02     \\
 Parkinson     & 0.95, 1.00     & 0.92, 0.98     & 0.90, 0.94     \\
 Bike sharing  & 0.94, 1.01     & 1.29, 1.51     & 0.82, 0.85     \\
 Inflation     & 0.98, 1.09     & 1.33, 1.58     & 1.35, 1.38     \\
 Obesity       & 0.94, 1.08 (1) & 2.10, 2.45     & 1.30, 1.86     \\
 Height        & 0.99, 1.14     & 1.53, 1.65     & 0.82, 1.37 (1) \\
 Death rate    & 0.99, 1.04     & 1.52, 1.62     & 2.24, 2.30     \\
 Sarcos        & 1.00, 1.04     & 0.99, 1.05 (1) & 1.11, 1.49     \\
\hline
\end{tabular} 
\end{table*}

\subsection{Effect of dataset size}\label{sec:result-data-size}
We now compare the sensitivity to dataset size,
by training the neural network architectures on reduced datasets.
The training datasets are cut to 50\% and 10\% of their original size. 
The same fraction of data is removed from each task, keeping the data balance from the original dataset.
Test sets are kept at full size. 
Training and hyperparameter searches are conducted in the same way as for the full data case.
The results are summarized in Table \ref{tab:error-datasize}.
Context-sensitive networks are on average slightly behind the others on all data sizes. 
A reduction in data size generally leads to a reduction in performance for all models,
but the learned-context architecture is less sensitive to this than the others.
Comparing the performance with 10\% of the data to that of the full dataset, three datasets stand out with an improved performance, namely Obesity, Height, and Sarcos.
These datasets have the highest number of data points and errors that are small in absolute value compared to the variance in the dataset.
This may make the variance due to the training procedure, studied in Table \ref{tab:rmse-variation}, play a greater role than the reduction in the dataset size for these cases.

\begin{table*}
\centering
\caption{Relative test errors for models trained on reduced datasets for
learned-context neural networks (LC),
context-sensitive neural networks (CS),
and last-layer neural networks (LL).
    Errors are normalized by the response standard deviation from Table \ref{tab:datasets}.
    }
\begin{tabular}{l|lll|lll|lll}
 \multicolumn{1}{c}{ } & \multicolumn{3}{c}{100\% training data} & \multicolumn{3}{c}{50\% training data} &  \multicolumn{3}{c}{10\% training data} \\
\hline
 Dataset       & LC            &   CS & LL            & LC            &   CS & LL            & LC            & CS            & LL            \\
\hline
 Frequency     & \textbf{0.29} & 0.37 & 0.33          & \textbf{0.31} & 0.36 & 0.37          & \textbf{0.49} & 0.87          & 0.53          \\
 Sine and line & 0.08          & 0.09 & \textbf{0.08} & 0.09          & 0.11 & \textbf{0.09} & \textbf{0.25} & 0.40          & 0.50          \\
 Schools       & \textbf{0.80} & 0.81 & 0.81          & \textbf{0.84} & 0.84 & 0.93          & \textbf{0.88} & 0.98          & 1.09          \\
 Parkinson     & 0.27          & 0.27 & \textbf{0.25} & 0.28          & 0.27 & \textbf{0.26} & 0.37          & \textbf{0.35} & 0.42          \\
 Bike sharing  & 0.30          & 0.44 & \textbf{0.25} & 0.37          & 0.56 & \textbf{0.28} & \textbf{0.54} & 0.74          & 0.70          \\
 Inflation     & \textbf{0.05} & 0.07 & 0.07          & \textbf{0.06} & 0.08 & 0.10          & \textbf{0.10} & 0.13          & 0.13          \\
 Obesity       & \textbf{0.02} & 0.05 & 0.04          & 0.04          & 0.05 & \textbf{0.02} & 0.04          & 0.10          & \textbf{0.02} \\
 Height        & 0.02          & 0.03 & \textbf{0.02} & 0.01          & 0.03 & \textbf{0.01} & 0.03          & 0.04          & \textbf{0.01} \\
 Death rate    & \textbf{0.10} & 0.15 & 0.22          & \textbf{0.12} & 0.17 & 0.19          & \textbf{0.20} & 0.30          & 0.58          \\
 Sarcos        & \textbf{0.12} & 0.12 & 0.13          & \textbf{0.12} & 0.12 & 0.16          & 0.13          & 0.12          & \textbf{0.10} \\
\hline
\end{tabular}
\label{tab:error-datasize}
\end{table*}

\subsection{Effect of task parameter dimension}\label{sec:results-dim-beta}
\begin{figure*}
    \centering
    \includegraphics[width=\columnwidth, page=1]{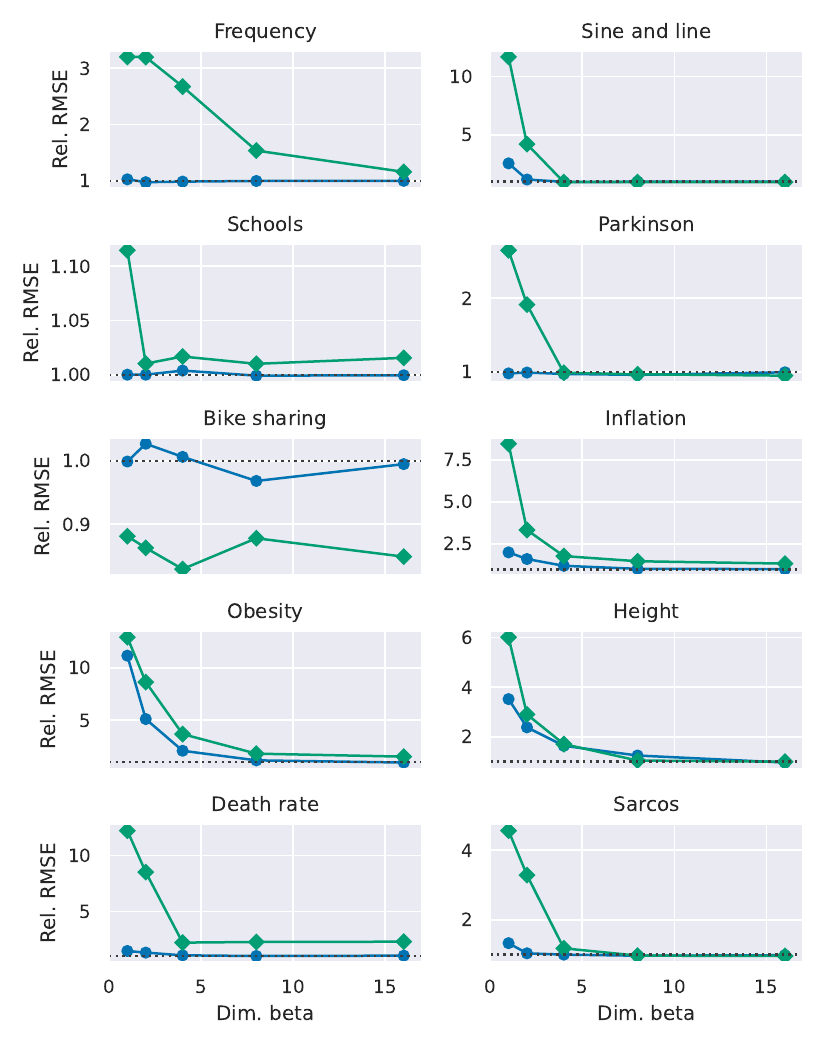} 
    \caption{
    Average test data error as a function of the number of task parameters. 
    Learned-context neural networks are given in blue circles and last-layer neural networks are in green diamonds. 
    Task parameter dimensions are set to 1, 2, 4, 8, and 16.
    Errors are normalized using the learned-context error from Table \ref{tab:error-base}. A dotted line at one marks the base performance.
    }
    \label{fig:dim-beta}
\end{figure*}

Section \ref{sec:universal-approximation} established theoretical differences in the number of task parameters required by learned-context networks and last-layer neural networks.
We now explore the practical impact of the task parameter dimension.
To this end,
learned-context networks and last-layer networks are trained on all datasets with different number of task parameters.
All other hyperparameters are fixed.
The models are trained on the full dataset. 
Figure \ref{fig:dim-beta} illustrates the results.
Additional task parameters generally improve the performance of both models on all datasets, up to a certain point.
There does not seem to be a significant downside to excessively large task parameter dimensions.
This means the hyperparameter searches will likely arrive at a larger than necessary value
unless additional model selection penalty terms are introduced.
Overall,
the learned-context neural networks achieve better performance with fewer task parameters,
but the gap is closed as the number increases.

As noted in Section \ref{sec:datasets},
the sine and line dataset is easily represented as regression on four nonlinear basis functions.
This is reflected by both models performing identically for four or more task parameters.
The frequency dataset does not map to linear regression in the same way. 
As a consequence, 
the last-layer neural network requires a significantly higher number of task parameters to achieve the same performance as the learned-context neural network.

\begin{figure*}
    \centering
    \includegraphics[width=\columnwidth, page=1]{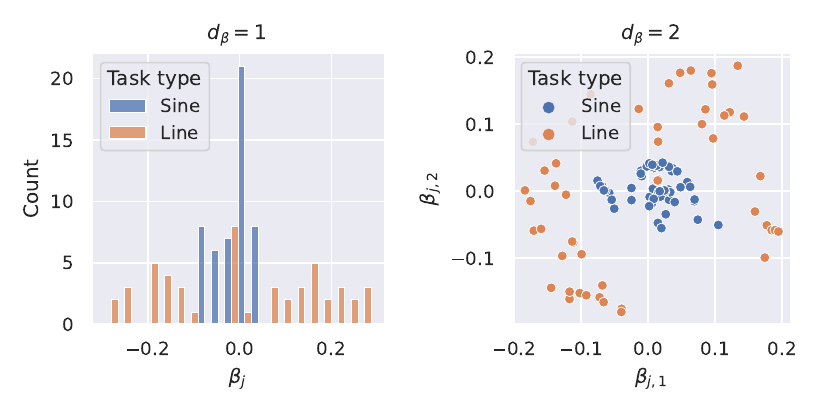} 
    \caption{
    Task parameters visualized for the Sine and line dataset.
    On the left is a model with one parameter for each task,
    illustrated as a histogram.
    On the right is a model with two parameters for each task,
    illustrated as a scatterplot.
    Both models are colored by the task type.
    On the left are the cases with one and two task parameters.
    The models are taken from the experiment given in Figure \ref{fig:dim-beta}.
    }
    \label{fig:sine-line-1d-2d}
\end{figure*}

\begin{figure*}
    \centering
    \includegraphics[width=\columnwidth, page=2]{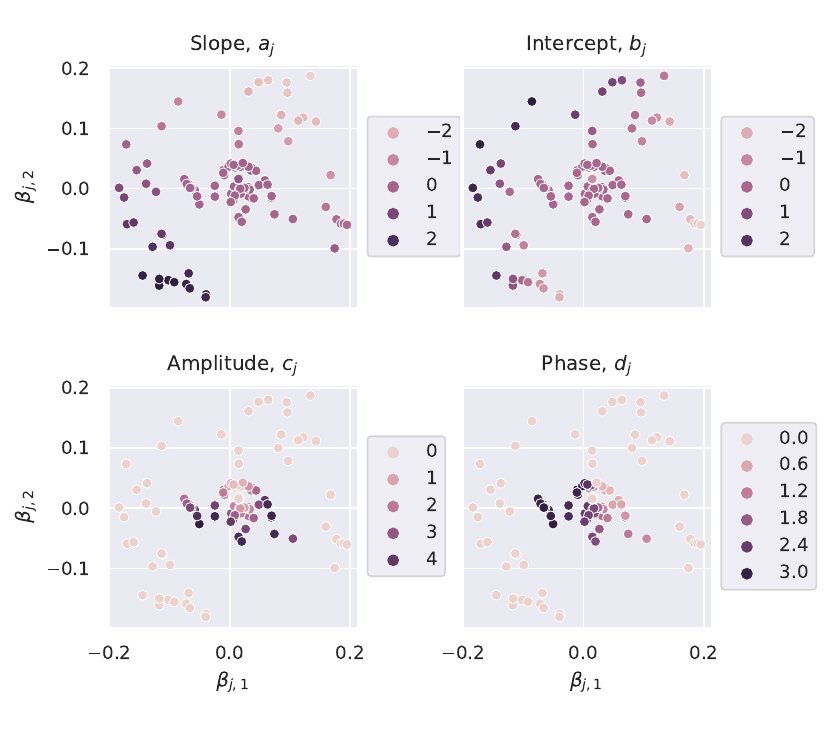} 
    \caption{
    Visualization of the relationship between the learned-context task parameter values and the underlying function parameters used to generate the data for sine and line case.
    The leaned-context neural network illustrated is the case with two dimensional task parameters illustrated in Figure \ref{fig:sine-line-1d-2d}.
    Each subplot is colored by one of the parameters used in the sine and line tasks.
    }
    \label{fig:sine-line-beta-vs-true-values}
\end{figure*}

Figure \ref{fig:dim-beta} illustrates the difference between the theoretical results from Section \ref{sec:universal-approximation} and learning in practice.
While a scalar parameter is sufficient to prove universal approximation for learned context networks, it is clear that for many problems it is beneficial to have a higher number of task parameters.
Figure \ref{fig:sine-line-1d-2d} illustrates how additional task parameters can help separate tasks that are fundamentally different, using the Sine and line dataset.
For both one and two task parameters we have that the sine tasks are close to the origin and the line tasks take parameters with larger magnitudes,
but the two task types are easier to separate in two dimensions.
The exception is a line task with both slope and intercept close to zero, which is placed almost at the origin along with sine tasks with small amplitude.
This is illustrated in Figure \ref{fig:sine-line-beta-vs-true-values}, 
which relates the learned-context task parameters to the parameters of the underlying functions that generated the data.
The two task parameters span out the underlying task properties in an interpretable manner. Note that the phase parameter of the sine tasks is periodic, which explains the apparent discontinuity.
Allowing for a larger task parameter space may be beneficial for certain learning problems, 
partially due to its ability to reduce negative inter-task interference when tasks are fundamentally different.

A benefit of a low-dimensional task parameter space is that it simplifies visualization and interpretation of the task parameters. Further examples of this are given in \ref{app:qualitative-study-task-parameters}, where the models for Inflation, Bike sharing, and Death rate datasets are studied in detail. The task parameters are given sensible interpretations related to the properties of the model domain.

A downside of a low-dimensional task parameter is,
 as discussed in Section \ref{sec:universal-approximation},
the unfortunate training dynamics that can be caused due to dying neurons.
In particular, the combination of a narrow first layer and a scalar task parameter can result in a lack of expressiveness. 
To address this, we used a minimum network width of 50 neurons, in combination with a residual architecture, to reduce the probability of being negatively affected by this phenomenon.
Figure \ref{fig:dim-beta} gives us confidence that this problem is not particularly dominant for the cases studied here.
For instance, we see that for many datasets the learned-context neural network is able to achieve almost identical performance for all task parameter dimensions, 
which indicates a reliable training procedure that is not heavily influenced by the randomness caused by dying neurons.
For other datasets, we see a stable increase in performance with the first few additional task parameters,
but this seems related to the ease of separating the tasks in a larger space, and not due to problems with the learning algorithm.

\subsection{Hold-out task performance} \label{sec:results-hold-out-task}
A key aspect differentiating the learned context from a fixed encoding, such as the context-sensitive neural networks, is that similar tasks can be assigned similar task parameter values. A desirable feature would be that the parameters capture latent properties of the tasks. For this to happen, the task parameter space must in some sense be well-behaved. A qualitative study, given in \ref{app:qualitative-study-task-parameters}, supports the hypothesis that learned-context neural networks facilitate such behavior, 
and the task parameters capture properties fundamental to the domain, as opposed to separating tasks by memorization.

This section attempts to quantify this behavior for the learned-context neural networks, by studying the viability of estimating parameters for a new task after the shared neural network parameters are trained and fixed. 
This is similar to the fine-tuning that motivated the factorization strategies of, among others, \citet{aghajanyan-etal-2021-intrinsic}, \citet{Mallya2018}, and \citet{Wang2023},
illustrated in Equation \ref{eq:intrinsic-dimension} and Equation \ref{eq:masked-network}.
Such fine-tuning resembles meta-learning. However, as discussed in Section \ref{sec:learning-paradigms}, meta-learning takes this generalization into account in the original optimization objective.  We only study these properties as a consideration \textit{after} the model has been trained using conventional multi-task learning. 
This experiment also resembles prompt-tuning for language models \citep{Lester2021power},
but differs in that the shared parameters are found by supervised learning on data from the same type of tasks as the hold-out tasks.

For these experiments, tasks are divided into two groups. 
The base group is the tasks that participate in the training of the shared parameters, by minimizing Equation \ref{eq:loss-train}.
The hold-out group consists of tasks that arrive after the shared parameters are fixed.
Hold-out tasks are considered one at a time.
The loss function for hold-out task $j$ is
\begin{align}
   \underset{\beta_j}{\min} \ 
   \frac{1}{s^2}\sum_{i=1}^{n_j} \left( y_{ij} - f(x_{ij}; \beta_j, \alpha) \right)^2 +
   \beta_j^\top D^{-1}\beta_j. \label{eq:loss-hold-out-task}
\end{align}
This is equivalent to a maximum a posteriori estimate of the task parameters, with a prior $\beta_j \sim \mathcal{N}(0, D)$, where $D$ is found from the distribution of task parameters in the base group. 
The log-likelihood term is scaled by $s^2$, which is the test error variance for the base group.

\begin{figure*}
    \centering
    \includegraphics[width=\columnwidth, page=1]{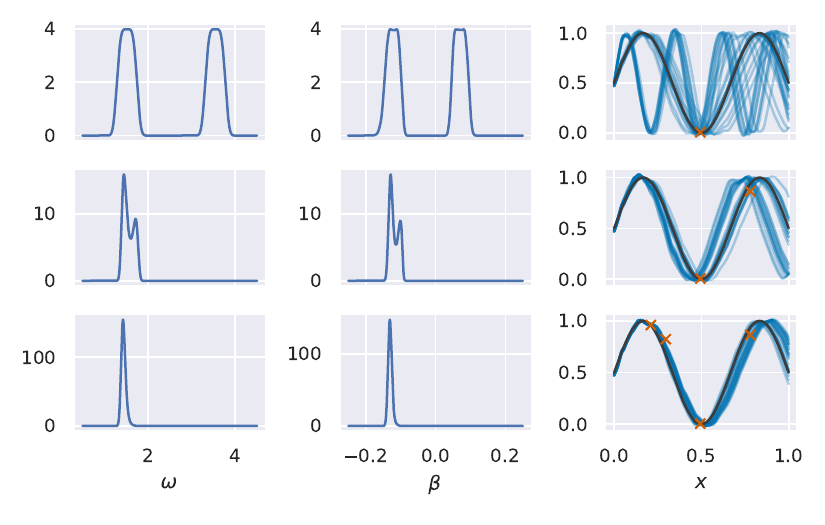} 
\caption{
    Task parameter likelihood functions for a hypothetical new task constructed for the Frequency dataset.
    The task has $\omega_j = 1.5$, and up to four data points.
    The left column is the likelihood of the true data-generating function, 
    which is a sine wave parametrized by its frequency.
    The middle column is the likelihood of the learned-context neural network, with one task parameter.
    The right column is the underlying function in black, the data points in orange,
    and the learned-context network evaluated with task parameters sampled from the likelihood in blue.
    The model is taken from the experiment in Section \ref{sec:results-dim-beta}.
    The rows show the likelihoods for a different number of data points.
    }
    \label{fig:likelihoods}
\end{figure*}

A critical part of parameter estimation is the shape of the likelihood function. 
Using the Frequency dataset, 
we compare the hold-out task likelihood with the true data-generating process, which in this case is known.
The result is given in Figure \ref{fig:likelihoods}.
Both parameter spaces display the same behavior for the different sets of data points. 
The modes of the likelihood develop predictably with additional data.
This is evidence of a well-behaved task parameter space,
and that the task parameters can capture properties that are fundamental to the problem, rather than just being a complex task encoding.
For the synthetic Frequency case, we were free to generate a sufficiently large number of tasks and data points for this to be possible.
The extent to which such a relationship between task parameters and underlying task properties can be identified is naturally dependent on the particular problem being studied.

To further investigate the hold-out task viability, an experiment is conducted on all datasets.
For each dataset, tasks are randomly divided into three approximately equal groups. 
One group is selected as the hold-out group, while the other two constitute the base group.
First, a learned-context neural network is trained using the base group tasks.
Task parameters for the hold-out group are then estimated one task at a time, with the shared parameters frozen.
The process is repeated using all groups as the hold-out group, and results are averaged over all iterations.
The loss in Equation \ref{eq:loss-hold-out-task} is minimized using the LIPO optimizer \citep{Malherbe2017}.

\begin{figure*}
    \centering
    \includegraphics[width=\columnwidth, page=1]{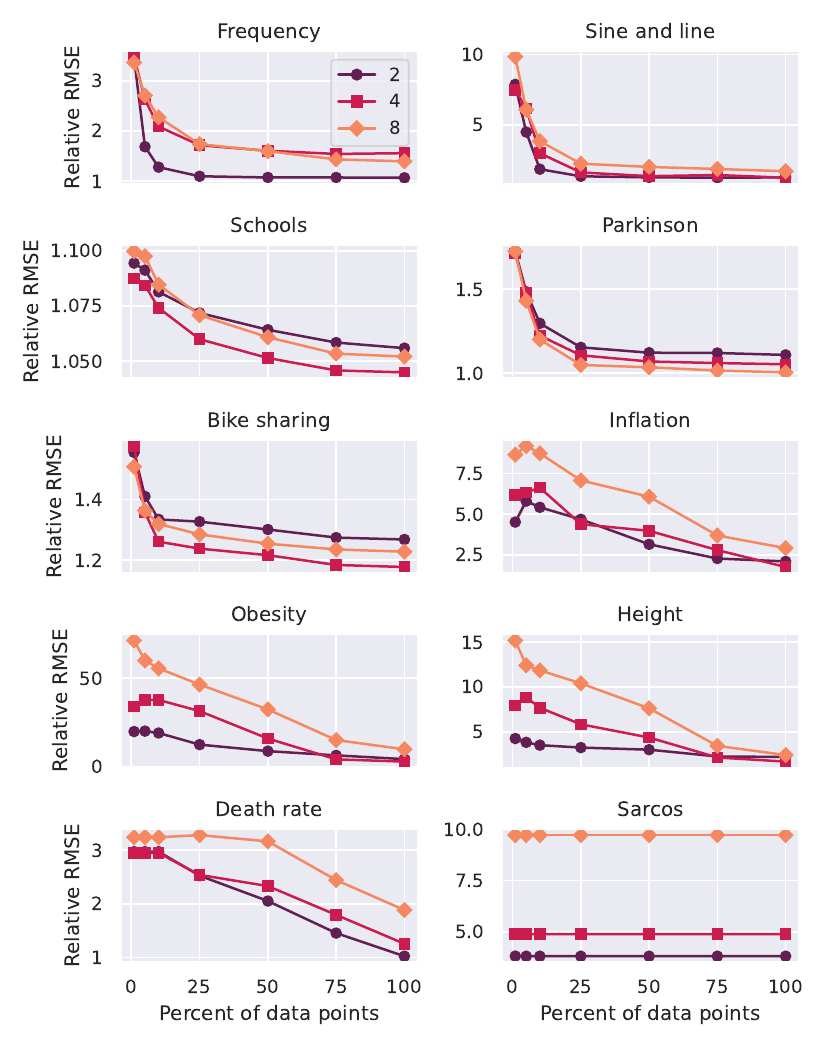} 
    \caption{
    Average test set errors for hold-out tasks.
    Errors are normalized using the learned-context network performance from Table \ref{tab:error-base}.
    Hold-out tasks are trained on 1\%, 5\%, 10\%, 25\%, 50\%, 75\%, and 100\% of the task data, 
    and evaluated on the full test set.
    Learned-context neural networks with 2, 4, and 8 task parameters are used. 
    }
    \label{fig:task-cv}
\end{figure*}

The results are summarized in Figure \ref{fig:task-cv} and Table \ref{tab:task-cv}.
The outcome of this experiment will naturally vary with the datasets.
In cases with few tasks or where tasks show a greater degree of uniqueness,
it will be hard to identify sensible parameters for the hold-out tasks. 
This can be seen clearly in the Sarcos dataset, which only has seven tasks, and the tasks are quite different.
For other cases, such as the Frequency dataset, there are many similar tasks, and the hold-out tasks are much more likely to resemble the base tasks.
Inspecting Table \ref{tab:task-cv}, the hold-out task performance is never able to match the baseline from Table \ref{tab:error-base}, 
but it is able to stay below a 20\% error increase for half of the datasets. 
In real scenarios where new tasks appear, this leads to a trade-off between retraining the full model for a potential performance gain, or just estimating the new task parameters to save time and resources.
As seen in Figure \ref{fig:task-cv}, it is beneficial to have more data, but the amount required for satisfying parameter estimation can vary greatly. 

\begin{table*}
\centering
\caption{
Mean test error for hold-out tasks when trained on 100\% of the training set for learned-context neural networks with 2, 4, and 8 task parameters.
Errors are normalized using the learned-context network performance from Table \ref{tab:error-base}.
}
\begin{tabular}{lrrr}
\hline
Dataset        &   $d_\beta$ = 2        &       $d_\beta$ = 4   & $d_\beta$ = 8 \\
\hline
 Frequency     & \bf{1.07} & 1.56 & 1.4  \\
 Sine and line & 1.28 & \bf{1.26} & 1.72 \\
 Schools       & 1.06 & \bf{1.04} & 1.05 \\
 Parkinson     & 1.11 & 1.05 & \bf{1.01} \\
 Bike sharing  & 1.27 & \bf{1.18} & 1.23 \\
 Inflation     & 2.08 & \bf{1.74} & 2.9  \\
 Obesity       & 4.17 & \bf{2.77} & 9.74 \\
 Height        & 2.18 & \bf{1.67} & 2.4  \\
 Death rate    & \bf{1.02} & 1.25 & 1.88 \\
 Sarcos        & \bf{3.81}  & 4.89 & 9.71 \\
\hline
\end{tabular}
\label{tab:task-cv}
\end{table*}

We note that when using fewer data points for the hold-out tasks, a lower task parameter dimension can be advantageous.
This is observed in the Obesity and Height datasets, where all three task parameter dimensions yield similar performance using the full dataset, but the smaller task parameter spaces become increasingly favored with fewer data points. 
Compare this to the results observed in Figure \ref{fig:dim-beta}, 
where it was found that an increasing task parameter dimension was favorable when all tasks were trained simultaneously. 
The optimal choice of task-parameter dimension is then up to the specific use case.

\section{Summary and discussion} \label{sec:discussion}
The theoretical and empirical results show that
the learned-context neural network is a favorable architecture for multi-task problems with similar tasks, and where tasks may have few data points. Its ability to learn low-dimensional and well-behaved task parameter spaces can be particularly advantageous for such problems.

Theoretically, scalar task parameters were shown to be sufficient for a learned-context neural network to achieve universal approximation of all tasks (Section \ref{sec:universal-approximation}). 
The contextual inputs facilitate interesting task adaptations even for small constructed networks 
(\ref{app:task-adaptation-example}).
The ability to adapt to different tasks naturally increases with the size of the neural network.
A potential downside to this flexibility is the possibility of overfitting to individual tasks, which counters the desirable properties of multi-task learning.
This puts emphasis on careful hyperparameter selection.

Experimentally it is seen that the ideal number of task parameters varies between problems, but the architecture can generally achieve a large degree of task adaptation with only a few task parameters (Section \ref{sec:results-dim-beta}). 
Increasing the task parameter dimension is observed to have a beneficial effect in cases where all tasks are trained simultaneously (Section \ref{sec:results-hold-out-task}). 
However, if the shared network model is to be used for new tasks, then a smaller parameter space may be preferable. 
The ideal task parameter dimension will likely have to be set by the practitioner in light of domain knowledge and the desired application of the model.
The architecture facilitates task parameters that capture latent properties of the tasks (Section \ref{sec:results-hold-out-task} and \ref{app:qualitative-study-task-parameters}), which can enable convenient workflows for updating and maintaining such models in practice.

Learned-context neural networks performed similarly to the benchmark architectures on the full datasets (Section \ref{sec:results-base}). 
On the reduced datasets the learned-context neural network had less performance deterioration than the others (Section \ref{sec:result-data-size}). 
All the neural networks have been built with only conventional fully connected layers, and our emphasis has been to study and compare the task adaptation,
rather than finding the most performant architecture for each individual dataset.
However, there is no reason to limit the use of learned-context parameters to this architecture.

Training of the learned-context networks appears to be robust, and comparable to the other architectures discussed (Section \ref{sec:results-base}).
The construction used to prove Theorem \ref{thm:ua} gives a motivation for initializing the task parameters to zero.
Randomly initialized task parameters would have a higher chance of getting stuck in local minima with ``task-encoding'' properties.
Zero initializing, on the other hand, encourages similar tasks to follow the same trajectories during training, which enables a grouping of similar tasks.
This likely promotes a well-behaved parameter space,
and reduces the chance that multiple regions represent the same phenomena.

\section{Conclusion} \label{sec:conclusion}
The learned-context neural network is a simple, but powerful, multi-task learning architecture.
Its properties make it well suited to problems with moderate amounts of data,
and in situations where a low-dimensional, well-behaved task parameter space is beneficial for the application and analysis of the model.
The task adaptation mechanism yields universal approximation capabilities with only a single task parameter. 
Empirical studies show that the ideal task parameter dimension varies with the domain and model application,
but the number of required task parameters is generally lower than that of comparable architectures.

\section*{Acknowledgements}
This work was supported by Solution Seeker Inc. and The Research Council of Norway.

\newpage

\appendix

\section{Task adaptation examples} \label{app:task-adaptation-example}

\begin{figure*}
    \centering
    \includegraphics[width=\columnwidth, page=1]{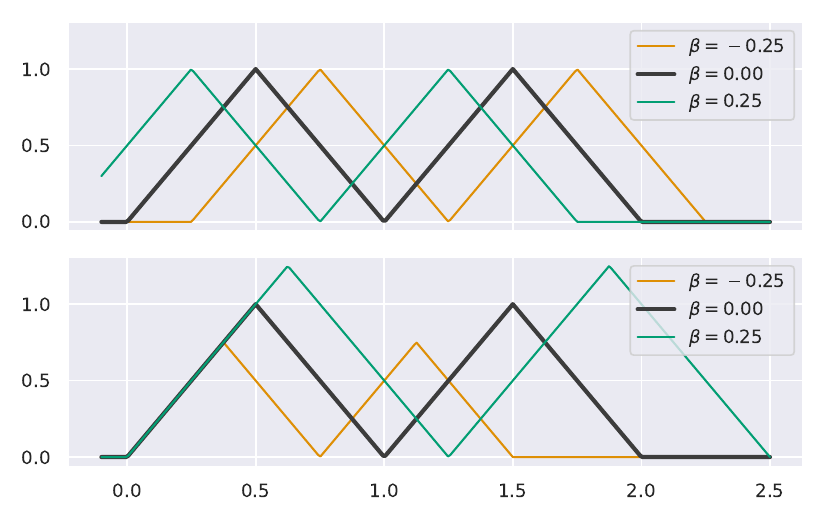} 
    \caption{
    Evaluation of the simple learned-context network described by Equations 
    \ref{eq:pyramid-1} to \ref{eq:pyramids-3}.
    The two examples only differ in their specification of the matrix 
    $L = \begin{bmatrix} L_1 & L_2 & L_3 & L_4 \end{bmatrix}^\top$. 
    The top figure is a translation case, with $L = \begin{bmatrix} 1 & 1 & 1 & 1 \end{bmatrix}^\top$,
    which is the first column of $W_1$. 
    The bottom figure is a dilation case, with $L = b_1$.
    Both have been evaluated with the task parameter, $\beta$, fixed at three different values.
    }
    \label{fig:pyramid}
\end{figure*}

We study a simple example to gain insight into how the learned context parameters can affect the network response.
The example is adapted from \citet{Telgarsky2016}.
A learned context network with three layers is constructed to linearly interpolate the points 
$(x, y) = (0, 0), (0.5, 1), (1, 0), (1.5, 1), (2, 0)$, 
when the task parameters are equal to zero.
Let the input be $z^\top = \begin{bmatrix} x & \beta \end{bmatrix}$,
with both $x$ and $\beta$ scalar.
The network parameters are given as
\begin{align}
 W_1 &= \begin{bmatrix}  1 & L_1 \\  1 & L_2 \\
 1 & L_3 \\
 1 & L_4 \\  \end{bmatrix},
 b_1 = -\frac{1}{2}\begin{bmatrix} 0 \\ 1 \\ 2 \\ 3 \end{bmatrix}, \label{eq:pyramid-1}
 \\
 W_2 &= \begin{bmatrix} 2 & -4 & 0 & 0 \\ 0 & 0 & 2 & -4 \end{bmatrix},
 b_2 =  \begin{bmatrix} 0 \\ 0 \end{bmatrix},  \label{eq:pyramids-2}
 \\
 W_3 &= \begin{bmatrix} 1 & 1 \end{bmatrix},
 b_3 = 0, \label{eq:pyramids-3}
\end{align}
and we use the ReLU activation function for the first two layers and the identity function for the last layer.
Briefly described, the first layer creates four identical unit slopes, starting at $0.0$, $0.5$, $1.0$, and $1.5$ when the task parameter, $\beta$, is zero.
The second layer adds pairs of these slopes together,
creating two triangle-shaped responses that are added together in the final layer.
The result is illustrated as the bold dark line in Figure \ref{fig:pyramid}.
We refer to this as the base shape.

We now explore the effect of different values for the  task parameter component $L$ from Equation \ref{eq:context-as-mixed-intercept}.
Starting with the simple case where $L = A$, which yields a translation of the base shape. 
This can be seen by manipulating Equation \ref{eq:nn-res-first-linear},
$z_{ij}^{(2)} = A x_{ij} + A \beta_j + b_1 = A \left( x_{ij} + \beta_j \right) + b_1$,
which is the expression for translation $y_{ij} = f(x_{ij} + \beta_j)$.
This is illustrated in the top part of Figure \ref{fig:pyramid}.
A different transform is obtained by setting $L = b_1$.
For this particular network, it yields dilation by a factor $1+\beta$.
This changes the base shape of the output to a dilated shape according to $(x, y) \mapsto ((1+\beta)x, (1+\beta)y)$.
This is illustrated in the bottom part of Figure \ref{fig:pyramid}.
The derivation of this result is given in \ref{app:dilation}.

These simple examples illustrate how the contextual inputs can produce different transformations by changing the first layer weights. While looking limiting at first glance, the ability to influence the first layer bias creates changes that propagate through the entire network, enabling powerful nonlinear transformations.

\subsection{Derivation of dilation effect} \label{app:dilation}
We show that setting $L = b_1$ in the network above creates a dilation effect. 
To simplify the analysis we take $\beta \in (-1, 1)$.
Let $z_{p}^{(2)} = x + L_p\beta + b_{1,p}$ be the $p$th element of the first hidden layer. Substituting $L_p = b_{1,p} = -(p-1)/2$, we get 
\begin{align*}
    g(z_{p}^{(2)}) = 
    \begin{cases}
    x - \frac{1+\beta}{2}(p-1) \ & \text{if} \ x \geq \frac{1+\beta}{2}(p-1)  \\
    0 \ & \text{otherwise}.
    \end{cases}
\end{align*}
Continuing the notation for the second hidden layer,
we get 
\begin{align*}
g(z_{1}^{(3)}) &= 
    \begin{cases}
    0,   \ & \text{if} \ x < 0, \\
    2x,  \ & \text{if} \ x \in \left [0, \frac{1+\beta}{2} \right ),  \\
    -2x + 2(1+\beta),  \ & \text{if} \ x \in \left [\frac{1+\beta}{2}, 1+\beta \right ),  \\
    0, \ & \text{if} \ x \geq 1+\beta,
    \end{cases} \\
g(z_{2}^{(3)}) &= 
    \begin{cases}
    0,   \ & \text{if} \ x < 1+\beta, \\
    2x - 2(1+\beta), \ & \text{if} \ x \in \left [1+\beta, 3\frac{1+\beta}{2} \right ),  \\
    -2x + 4(1+\beta),  \ & \text{if} \ x \in \left [3\frac{1+\beta}{2}, 2(1+\beta) \right ),  \\
    0, \ & \text{if} \ x \geq 2(1+\beta),
    \end{cases}
\end{align*}
The output is then given as $y = g(z_{1}^{(3)}) + g(z_{2}^{(3)})$,
which is a piecewise linear function interpolating the points
$(x, y) =  (0, 0), (0.5(1+\beta), 1+\beta), (1+\beta, 0), (1.5(1+\beta), 1+\beta), (2(1+\beta), 0)$. 
This is equivalent to a dilation of both $x$ and $y$ with a factor $1+\beta$.

\section{Optimizer and learning rate schedule} \label{app:optimizer}

All neural networks are implemented and trained with PyTorch \citep{Paszke2019}.
They are trained on a single GPU.
Optimization is done by stochastic gradient descent with momentum \citep{Bottou2018} and a learning rate scheduler.

The learning rate scheduler has two stages.
Starting at $1^{-8}$, it is first increased linearly during a warm-up stage \citep{Nakamura2021, Arpit2019} until it reaches peak learning rate.

The second stage is to train the model over several epochs until the loss converges\citep{chee18a, Lang2019}, at which point the learning rate is reduced by half. 
This is repeated until the learning rate is reduced back down below $1^{-8}$, the maximum number of epochs is reached, or the loss is sufficiently close to zero.
Loss convergence is determined by inspecting a window of the last epoch losses. 
A linear regression model with slope and intercept is fitted to the loss values.
This is compared to a model with intercept only, using a likelihood ratio test \citep{Wilks1938}. 
Convergence is flagged if the test is above 0.51, which is an aggressive threshold.
The test is implemented in Statsmodels \citep{Seabold2010}.
The new learning rate is kept for a minimum number of epochs equal to the window size.

The number of epochs and data batches varies with dataset size. 
For data sets with less than 100 000 data points, we use 10 000 epochs of two batches,
otherwise, we use 1000 epochs of 20 batches.
The warm-up stage is equal to 10\% of the maximum allowed epochs, and the loss convergence is found with a window size equal to 1\% of the epochs.
Peak learning rate and momentum are treated as hyperparameters.

Neural network parameters are initialized by the He initializer \citep{He2015}.
Task parameters are initialized to zero.

Linear mixed effect models are implemented and optimized with Statsmodels \citep{Seabold2010}.

All data is transformed to approximately unit scale before training and evaluating models.
Still, all figures and errors are given in the original units, unless stated otherwise.

\section{Hyperparameters}\label{app:hps}

All three neural network architectures require the number of residual blocks, hidden layer size, and network parameter regularization factor as hyperparameters.
They also required the peak optimizer learning rate.
Additionally, learned-context networks and last-layer neural networks require the number of task parameters and the task parameter regularization factor.

Hyperparameters are optimized using the training data in Table \ref{tab:datasets}. The training data is split into two parts, using one part for training and one part for validation during the search.
A final model is then trained on all training data using the hyperparameter configuration with the best validation error.
Optimization is done by a variant of the LIPO solver \citep{Malherbe2017} implemented in dlib \citep{King2009}.
The search runs for 25 iterations in all cases.

The range of values explored in the searches varies by dataset. Details are summarized in Table \ref{tab:hps-space}.
To limit the number of hyperparameters, 
momentum is fixed to 0.7 and the number of residual blocks is fixed to two for all datasets and architectures.
These values are found as reasonable choices for most cases.

Due to a large number of experiments and hyperparameter searches,
we observe that some searches arrive at an optimal learning rate that is too high to be used in training the final model. 
It can be due to randomness in weight initialization and batch samples that allowed a high learning rate to succeed during the trials, 
only to diverge during final training. 
To address this, we multiply the peak learning rate by 0.9 and retry the training if the loss diverges.

\begin{table*}
\centering
\caption{
Summary of hyperparameter search space. 
The number of task parameters is limited to the number of tasks $m$ in the dataset, up to a maximum of 25.}
\label{tab:hps-space}
\begin{tabular}{lll}
\hline
 Name                           & Min           & Max           \\
\hline
Peak learning rate              &   $10^{-4}$   & 1.5           \\
Hidden layer size               &            50 & 500           \\
Shared parameter regularization &   $10^{-15}$  & $10^{-5}$       \\
Task parameter regularization   &   $10^{-15}$  & $10^{-3}$       \\
Number of task parameters       &            1  & min(25, $m$)  \\
\hline
\end{tabular} 
\end{table*}

Full hyperparameter searches are conducted for the experiments in Section \ref{sec:results-base}
and Section \ref{sec:result-data-size}.
For the experiment in Section \ref{sec:results-dim-beta},
the hyperparameters are copied from the final models in Section \ref{sec:results-base}.
For the experiment in Section \ref{sec:results-hold-out-task},
the hyperparameters are copied from the experiment using 50\% training data in Section \ref{sec:result-data-size}.

\section{Qualitative study of task parameters}\label{app:qualitative-study-task-parameters}

This section provides additional visualizations of the learned-context neural networks trained in Section \ref{sec:results-empirical}.
The intention is to give further insight into the qualitative behavior of the task parameters. 
We study the Inflation, Bike sharing, and Death rate datasets because they have a low dimensional input space that is convenient to visualize.
The models with a scalar task parameter from Section \ref{sec:results-dim-beta} are used for all visualizations.

\begin{figure*}
    \centering
    \includegraphics[width=\columnwidth, page=1]{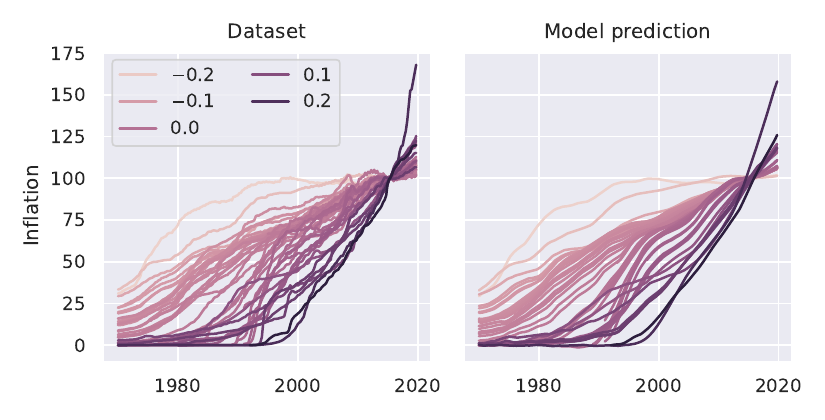} 
    \caption{
    Inflation dataset. 
    The left plot is the data points, where each line is the normalized inflation rate for a country (recall that each country is a task). 
    The right plot is the fitted models evaluated at the data points.
    In both plots, the lines are colored by the value of the task parameter for that country.
    }
    \label{fig:app-beta-inflation}
\end{figure*}

Figure \ref{fig:app-beta-inflation} illustrates the Inflation data and the fitted model.
The model appears as a smoothed version of the data.
The task parameters range from -0.2 to 0.2. 
Higher task parameter values seem to represent countries where the increase began in the 1990s, and lower values represent countries where the increase was well underway in the 1970s.
The transition between these two categories is highly non-linear.
For a task parameter equal to -0.2, the curve is initially steep and flattens towards the end. The curve then gradually transitions to an almost linear trend for values around -0.1.
For even higher values, there curve starts with a flat phase that eventually ramps up, with the start time increasing along with the task parameter.

\begin{figure*}
    \centering
    \includegraphics[width=\columnwidth, page=1]{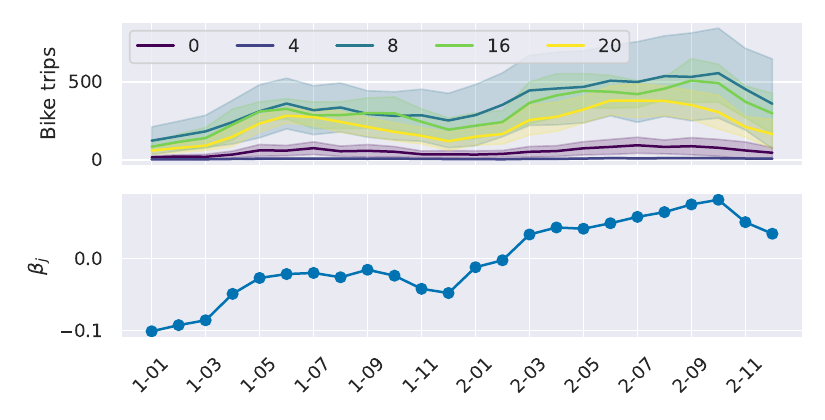} 
    \caption{
    Bike sharing dataset. 
  	The top plot is the average number of trips at different times of day for each month,  with the standard deviations given as transparent bands. 
    This plot only includes data from workdays.
 The bottom plot is the identified task parameter for each month.
    }
    \label{fig:app-beta-bikesharing}
\end{figure*}

The bike-sharing dataset relates the number of bike trips to weather, calendar features, and time of day, over two years. Each month is a task, yielding 24 tasks. Figure \ref{fig:app-beta-bikesharing} illustrates the average number of bike trips and the task parameters for each month. There is an increasing number of trips over the two years, combined with a seasonal effect of more activity during the summer months. The increase in activity in the second year may be due to an increase in the capacity or popularity of the system. The task parameters nicely follow the average trips taken during peak hours. A possible interpretation is that the task parameters capture some latent state of the system, such as the number of bikes being distributed. We emphasize that while the parameters are visualized as a connected line, they are \textit{not} regularized to follow a time-dependent pattern. All parameters are assumed to be independent and centered around zero.  However, a time-dependent regularization between the task parameters could be considered if the model were to be used in practice.

\begin{figure*}
    \centering
    \includegraphics[width=\columnwidth, page=1]{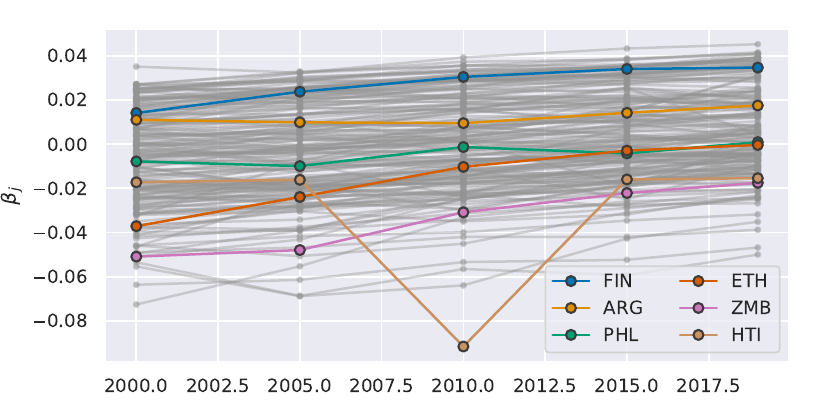} 
    \caption{
    Death rate dataset. 
    Visualization of task parameters. 
    Each combination of country and year is an independent task in the model formulation, but tasks from the same country are connected by a line. 
    Six countries are highlighted in colors.
    }
    \label{fig:app-beta-deathrate-1}
\end{figure*}

\begin{figure*}
    \centering
    \includegraphics[width=\columnwidth, page=2]{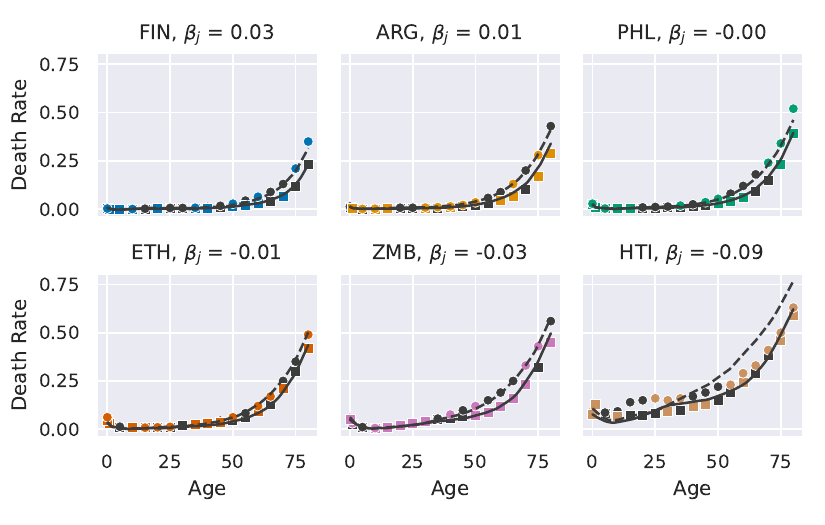} 
    \caption{
    Death rate dataset. Data and fitted models for six countries for the year 2010.
    The country and corresponding task parameters for this year are given in the titles.
    Data is given in black and colored markers, where black is the training data and colored is the test data. 
    Circle markers are used for males and square markers for females.
    Fitted models are given in dashed lines for males and solid lines for females.
    }
    \label{fig:app-beta-deathrate-2}
\end{figure*}

The death rate dataset studies the rate of death at different ages, as a function of country, year, and sex. Age is grouped in intervals of five years, and the maximum age included is 80.
All ages above 80 are grouped together with a death rate of one, which has been removed from the dataset in this study.
The data is given in five different years, 2000, 2005, 2010, 2015, and 2019. Each country and year is modeled as an individual task, making each country associated with five tasks. 
This relationship is \textit{not known} by the model. Figure \ref{fig:app-beta-deathrate-1} illustrates the task parameters, with six countries highlighted. The year 2010 is investigated in detail in Figure \ref{fig:app-beta-deathrate-2}, which illustrates the data and fitted models for the highlighted countries in Figure \ref{fig:app-beta-deathrate-1}.  It seems that lower task parameter values relate to higher death rates in general. For most countries the task parameters seem to be incrementally increased with the years, indicating a decrease in death rates for ages below 80. For instance, task parameters for Ethiopia (ETH) show a steady increase over time. This correlates with the increase in life expectancy at birth observed over the same period \citep{Ethiopia2022}.

Haiti (HTI) 2010 stands out in Figure \ref{fig:app-beta-deathrate-1}. 
This is due to the devastating earthquake of January 2010, which had a large death toll \citep{Cavallo2010}.
In Figure \ref{fig:app-beta-deathrate-2} we see that this leads to a curve shape that is quite unique, with an elevated death rate across all ages.
In this case, the other tasks fail to supply the information missing from the training data, and the resulting model overshoots on the test data.
For the other tasks, the gaps left by the test data are covered nicely by related tasks. 
For instance, the female death rate in Zambia (ZMB) is almost perfectly captured with only six data points available for training, of which only one represents ages 50 and above.

We emphasize that the task parameters are not forced into the relationships seen in Figure \ref{fig:app-beta-inflation},
Figure \ref{fig:app-beta-bikesharing}, and Figure \ref{fig:app-beta-deathrate-1}.
The continuous and interpretable task parameters are only motivated by regularization towards zero. 
The discovered structures are due to the information in the training data.

\bibliographystyle{elsarticle-harv} 
\bibliography{references}

\end{document}